\def\year{2020}\relax
\documentclass[letterpaper]{article} 
\usepackage{aaai20}  
\usepackage{times}  
\usepackage{helvet} 
\usepackage{courier}  
\usepackage[hyphens]{url}  
\usepackage{graphicx} 
\usepackage{subfig}
\usepackage{booktabs}
\usepackage{amsmath}
\usepackage{amssymb}
\usepackage{algorithm}
\usepackage{algpseudocode}

\def\ie{\emph{i.e. }}
\def\vs{\emph{v.s. }}
\def\eg{\emph{e.g. }}

\def\etal{\emph{et al. }}

\usepackage{graphicx}  
\title{Ultrafast Photorealistic Style Transfer via Neural Architecture Search}
\author{Jie An,\thanks{This work was done when Jie An worked as an intern at Big Data Lab of Baidu Research.}\thanks{Equal contribution.}\textsuperscript{\rm 1} Haoyi Xiong,\footnotemark[2]\textsuperscript{\rm 2} Jun Huan\textsuperscript{\rm 3} and Jiebo Luo\textsuperscript{\rm 1}\\ 
\textsuperscript{\rm 1}University of Rochester, \textsuperscript{\rm 2}Baidu Research, \textsuperscript{\rm 3}StylingAI Inc.\\ 
\{jan6, jluo\}@cs.rochester.edu, xionghaoyi@baidu.com, lukehuan@shenshangtech.com 
}
 
\begin{document}

\maketitle

\begin{figure*}[t]
    \centering
    \includegraphics[width=0.85\textwidth]{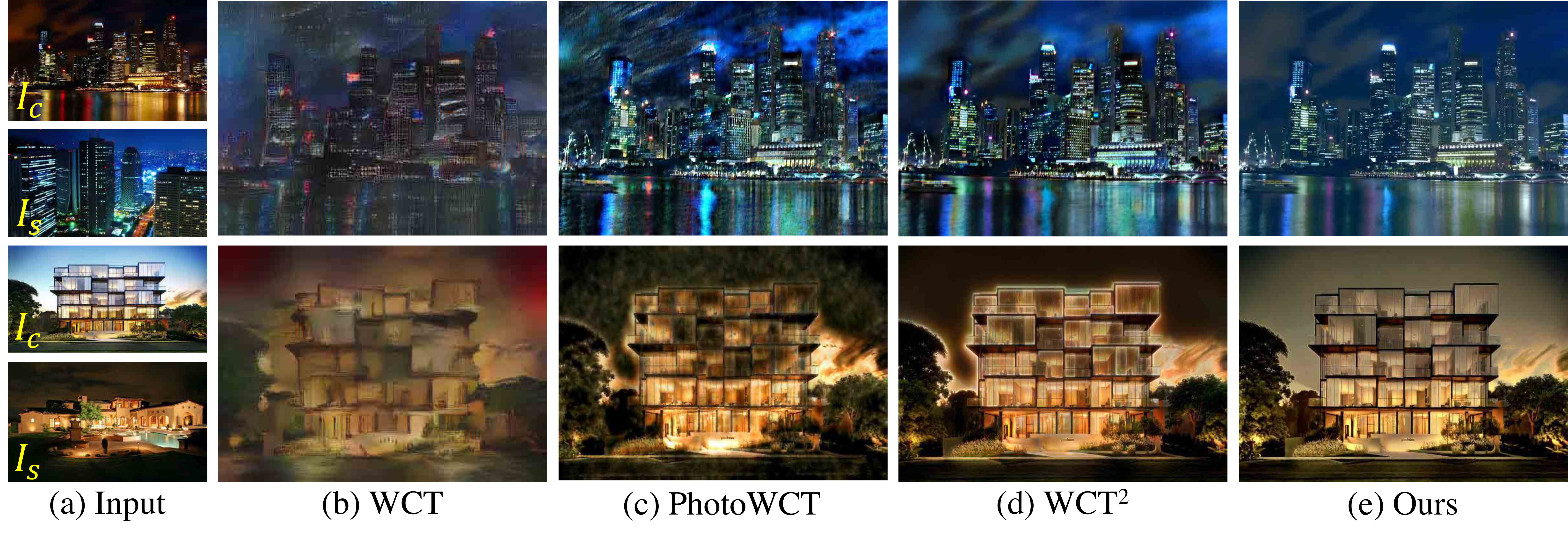}
    \caption{\textbf{Photorealistic style transfer results.} Given (a) an input pair ($I_c$: content, $I_s$: style), we show results of (b) WCT~\cite{li2017universal}, (c) PhotoWCT~\cite{li2018closed}, (d) WCT$^2$~\cite{yoo2019photorealistic}, and (e) our method. Every result is produced \emph{without} the assist of region masks and/or post-processing for a fair comparison. While the compared methods produce significant spatial distortions, the proposed approach achieves better style transfer results in terms of fine detail preservation and photorealism.}
    \label{fig:1}
\end{figure*}

\begin{abstract}
    The key challenge in photorealistic style transfer is that an algorithm should faithfully transfer the style of a reference photo to a content photo while the generated image should look like one captured by a camera. Although several photorealistic style transfer algorithms have been proposed, they need to rely on post- and/or pre-processing to make the generated images look photorealistic. If we disable the additional processing, these algorithms would fail to produce plausible photorealistic stylization in terms of detail preservation and photorealism. In this work, we propose an effective solution to these issues. Our method consists of a construction step (C-step) to build a photorealistic stylization network and a pruning step (P-step) for acceleration. In the C-step, we propose a dense auto-encoder named PhotoNet based on a carefully designed pre-analysis. PhotoNet integrates a feature aggregation module (BFA) and instance normalized skip links (INSL). To generate faithful stylization, we introduce multiple style transfer modules in the decoder and INSLs. PhotoNet significantly outperforms existing algorithms in terms of both efficiency and effectiveness. In the P-step, we adopt a neural architecture search method to accelerate PhotoNet. We propose an automatic network pruning framework in the manner of teacher-student learning for photorealistic stylization. The network architecture named PhotoNAS resulted from the search achieves significant acceleration over PhotoNet while keeping the stylization effects almost intact. We conduct extensive experiments on both image and video transfer. The results show that our method can produce favorable results while achieving 20-30 times acceleration in comparison with the existing state-of-the-art approaches. It is worth noting that the proposed algorithm accomplishes better performance without any pre- or post-processing.
\end{abstract}

\section{Introduction}
Photorealistic style transfer is an image editing task aims at changing the style of a photo to a given reference. To be photorealistic, the produced image should preserve spatial details of the input and looks like a photo captured by a camera. For example, in Fig.~\ref{fig:1}, we transfer the night view photo from a warm color to cold while in the other example, a day-time photo is changed to a night-time one. In these examples, the scene of the input content keeps intact in the produced result. Unfortunately, artistic style transfer methods~\cite{gatys2015neural,Gatys2016,johnson2016perceptual,ulyanov2016texture,li2017universal,huang2017arbitrary,sheng2018avatar,li2018learning} generally distort fine details (lines, shapes, borders) in images, which is necessary for producing art flavors in artistic scenarios but is not favored in photorealistic stylization. We illustrate the failure of artistic methods in photorealistic stylization cases with the example of WCT in Fig.~\ref{fig:1} (b). More failure cases are available in supplementary materials.

Based on Gatys~\etal\cite{Gatys2016}, Luan~\etal\cite{luan2017deep} introduce a photorealistic loss term and adopts an optimization method to make style transfer. However, solving the optimization problem is time/computation consuming. To address this issue, Li~\etal propose PhotoWCT~\cite{li2018closed} which uses a feed-forward network to make style transfer. Although PhotoWCT applied multi-level stylization and uses unpooling operator as a replacement of upsampling to enhance the detail preservation of the network, the produced results still suffer from distortions as demonstrated in Fig.~\ref{fig:1} (c). To overcome the remaining artifacts, they have to introduce close-formed post-processing and regional masks (if available) to regulate the spatial affinity of the image. However, such post-processing is computationally expensive and causes the result over-smoothed. Recently, Yoo~\etal~\cite{yoo2019photorealistic} proposed Wavelet Corrected Transfer (WCT$^2$) aims at eliminating post-processing steps while preserving fine details in transferred photos. Although using wavelets can increase the fidelity of signal recovery, WCT$^2$ also need to rely on region masks of content and reference style photos to perform style transfer. If such region masks are disabled, as shown in Fig.~\ref{fig:1} (d), the result of WCT$^2$ shows significant distortions. Since such region masks are hard to acquire for arbitrary photos (generally have to train specific networks to segment input photos and manually fine-tune the segmented results), the practical usage of PhotoWCT and WCT$^2$ is limited.

Regarding the network architecture, PhotoWCT and WCT$^2$ both adopt the same symmetric auto-encoder but use different downsampling and upsampling modules. However, general network architectures specifically designed for photorealistic style transfer have not been well investigated. This work fills this gap. Specifically, our algorithm consists of a network construction step (C-step) that introduces a highly-effective auto-encoder for photorealistic stylization, and a pruning step (P-step) is adopted in the following to compress the auto-encoder for acceleration. In C-step, we firstly conduct a carefully designed pre-analysis and introduce two architectural modules named \emph{Bottleneck Feature Aggregation (BFA)} and \emph{Instance Normalized Skip Link (INSL)} based on analyzed results. BFA, motivated by~\cite{yu2018deep,zhao2017pspnet}, employs multi-resolution deep features to improve photorealistic stylization effects. INSL is the combination of the Skip Connection (SC) originated from U-Net~\cite{ronneberger2015u} and the Instance Normalization~\cite{ulyanov1607instance}. INSL achieves high fidelity information recovery while avoiding ``short circuit'' phenomenons occurred when using SCs. Based on these modules, we constructed an asymmetric auto-encoder (named \textbf{PhotoNet}) with BFA and densely placed INSLs. Thanks to the proposed modules, our PhotoNet outperforms DPST~\cite{luan2017deep}, PhotoWCT and WCT$^2$ in terms of fine detail preservation. In P-step, we propose a Neural Architecture Search framework in a manner of teacher-student learning (namely StyleNAS). Here PhotoNet is the maximum architecture in our search space of NAS, where an evolution algorithm~\cite{kim2017nemo} is adopted to iterative prune removable operators (any operator except the VGG encoder and minimal basic operators to form a decoder) in PhotoNet. In each loop of the architecture search, we first mutate 20 new architectures. Each architecture contains a pre-trained VGG-19~\cite{simonyan2014very} as the encoder and the decoder is trained to reconstruct images. A validation process is adapted after training, where the performance of each architecture is evaluated by its similarity to the result of the oracle (\ie, PhotoNet). To compress network architectures, we additionally introduce a network complexity loss to penalize time-consuming networks and finally get a bunch of highly-efficient and effective networks for photorealistic style transfer. We pick up one of them (named \textbf{PhotoNAS}) for comparison in this paper and more searched architectures and its results are available in supplementary materials.

Our contributions are two-fold. For photorealistic style transfer, PhotoNet/PhotoNAS are the first networks that  \textit{do not require any post-processing or region mask assistance}. PhotoNAS is surprisingly simple and highly-efficient with \textit{$356\times$ speed up over PhotoWCT and $24\times$ over WCT$^2$ on $1024\times512$ photos}. PhotoNAS quantitatively outperforms the  state-of-the-art methods in terms of SSIM-Edge, SSIM-Whole, Gram Loss, and user preference percentage. Further experiments on video style transfer demonstrate its ability to stylize and produce stable videos without any specific modification. On the other hand, for Automatic Machine Learning (AutoML) and NAS, \textit{our algorithm is the first that successfully adopts NAS to design style transfer networks for photorealistic rendering}, which expands the application area of NAS to the style transfer area.

\section{Related Work}
\noindent\textbf{Style Transfer.}
Significant efforts have been made to image style transfer in the area of computer vision. Prior to the adoption of deep neural networks, several classical models based on stroke rendering~\cite{hertzmann1998painterly}, image analogy~\cite{hertzmann2001image,shih2013data,shih2014style,frigo2016split,liao2017visual}, or image filtering~\cite{winnemoller2006real} have been proposed to make a trade-off between quality, generalization, and efficiency for style transfer.

Gatys~\etal~\cite{gatys2015neural,Gatys2016} first proposed to model the style transfer as an optimization problem minimizing deep features and their Gram matrices of neural networks, while these networks were designed to work well with artistic styles only.
In photo style transfer scenarios, neural network approaches~\cite{luan2017deep,li2018closed} have been proposed to enable style transfer for photorealistic styles. These methods either introduce smoothness-based loss term~\cite{luan2017deep} or utilize post-processing to smooth the transferred images~\cite{li2018closed}, which inevitably decreased fine details of images and increased time-consumption significantly. Recently, Yoo~\etal~\cite{yoo2019photorealistic} proposed WCT$^2$, which allows transferring photorealistic styles without inefficient post-processing. However, WCT$^2$ has to work with the assist of region masks, which are hard to acquire and thereby limit its practical applications.

\noindent\textbf{Image-to-image Translation.}
In addition to style transfer, photorealistic stylization has also been studied in image-to-image translation~\cite{isola2017image,wang2018high,liu2016coupled,taigman2016unsupervised,shrivastava2017learning,liu2017unsupervised,zhu2017unpaired,huang2018multimodal}.
The major difference between photorealistic style transfer and image-to-image translation is that photorealistic style transfer does not require paired training data (i.e., pre-transfer and post-transfer images). Of course, image-to-image translation can solve even more complicated task such as the  man-to-woman and cat-to-dog adaption problems.
\begin{figure}[t]
    \centering
    \includegraphics[width=\linewidth]{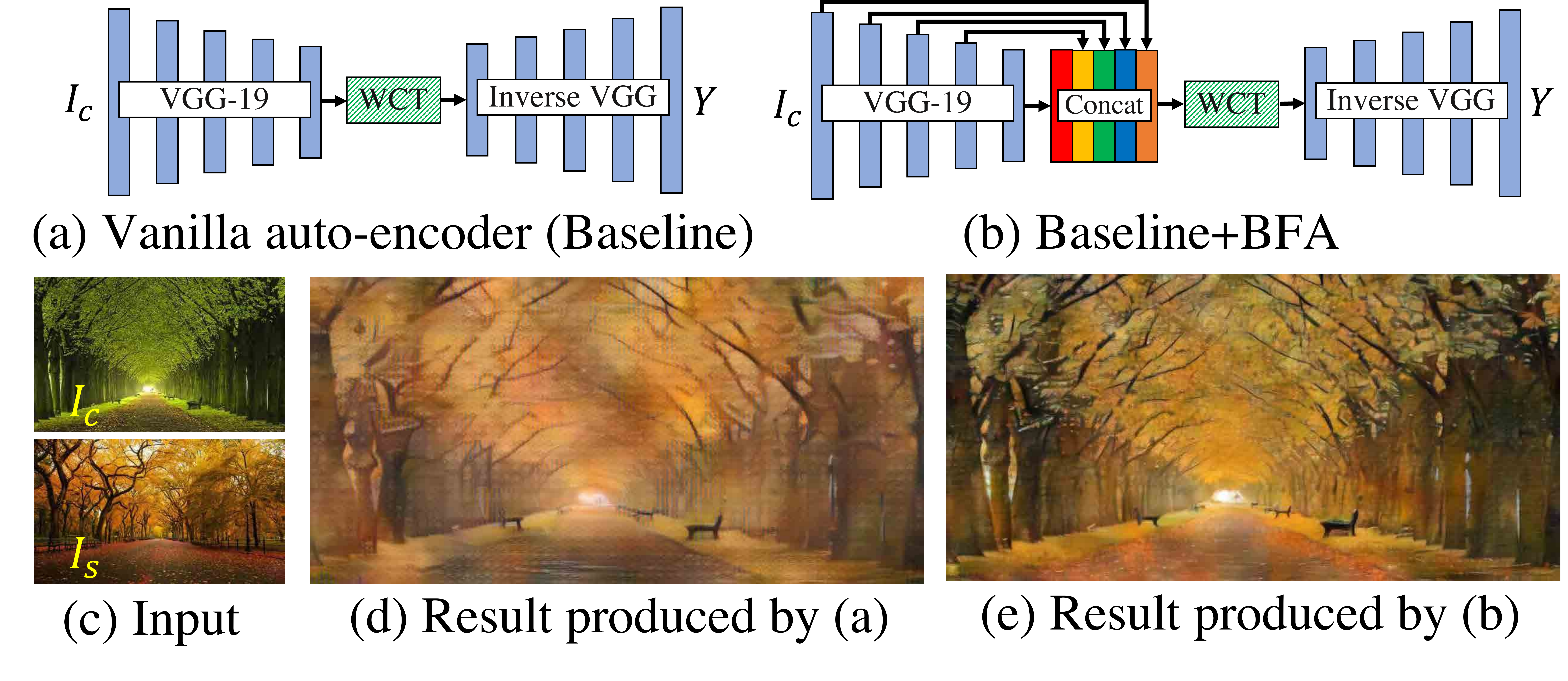}
    \caption{\textbf{Comparison between auto-encoders \emph{with} and \emph{without} BFA.} (a) is the vanilla auto-encoder with WCT as the transfer module placed in the bottleneck, which is used as the baseline. (b) is the auto-encoder equipped with BFA module. (c) is the input content ($I_c$) and style ($I_s$) images. (d) and (e) are results produced by (a) and (b) respectively. Trees in (e) contain comparably more detailed branches and leaves.}
    \label{fig:2}
\end{figure}

\noindent\textbf{Discussion.}
The work most relevant to our study includes WCT, PhotoWCT, and WCT$^2$. WCT has been used for artistic stylization and the last two ones are for photorealistic stylization. Compared with PhotoWCT, the proposed method can avoid time-consuming post-processing and multi-round stylization while ensuring the effectiveness of style transfer. The main difference between our approach and WCT$^2$ is that the proposed algorithm allows transferring photo styles without any assist of region masks acquired by segmenting content and style inputs. Compared with PhotoWCT and WCT$^2$, the results produced by our method has considerably higher sharpness, fewer distortions and a remarkable reduction of computational cost.

\section{Pre-analysis}
To design effective modules/networks for photorealistic style transfer, we start with conducting a pre-analysis on architectural factors may influence stylization effect to propose useful network modules for the enhancement of stylization performance. We adopt a vanilla symmetrical auto-encoder as the baseline. For each studied module, we will compare its transfer results with the baseline in terms of visual effects and photorealism. More analyzed results are available in supplementary materials.

\begin{figure}[t]
    \centering
    \includegraphics[width=\linewidth]{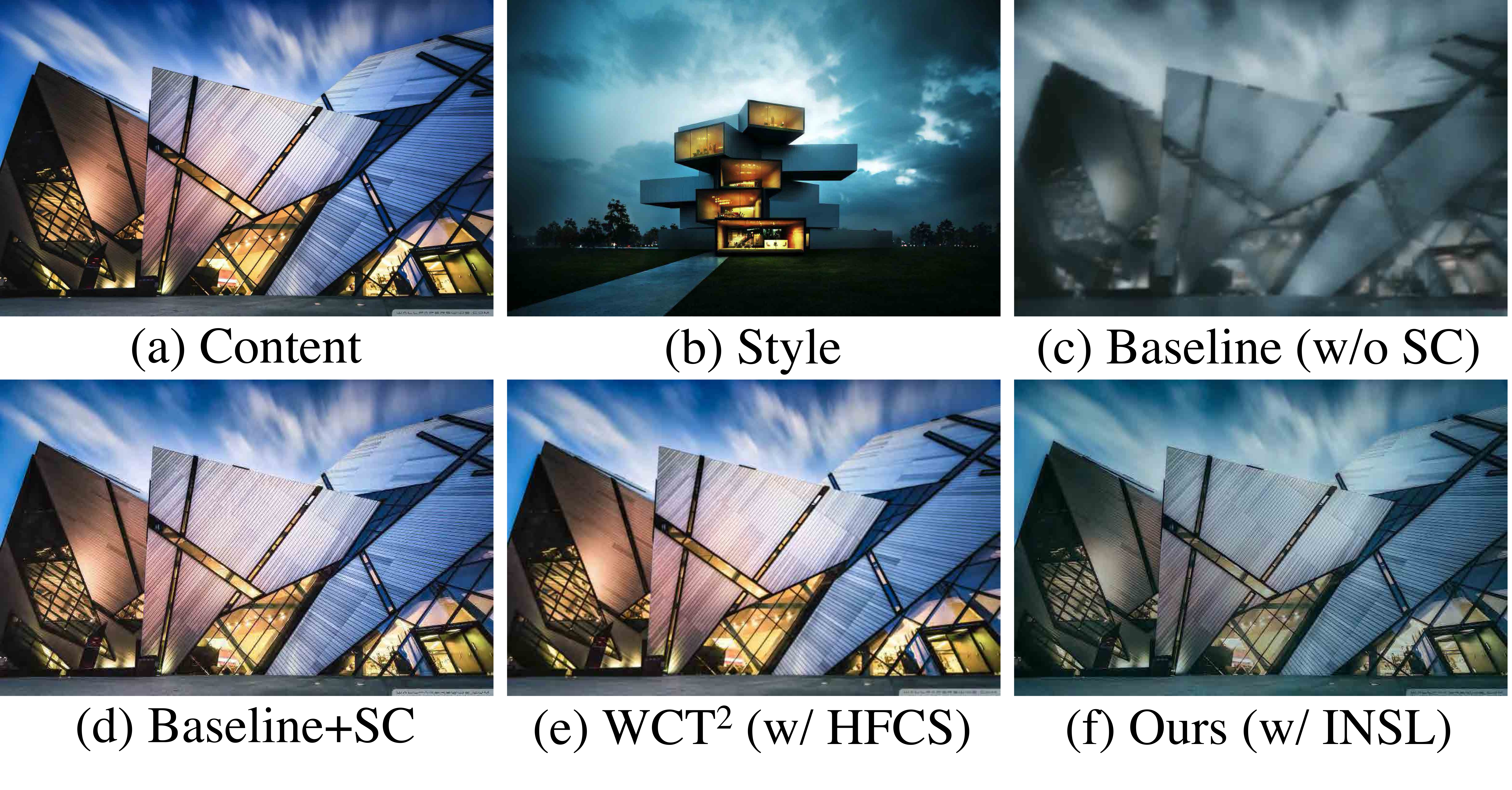}
    \caption{\textbf{Comparison of SC, HFCS and INSL.} SC (d) causes the ``short circuit'' issue that removes stylization effects of the baseline (c). Similar failure case also exists in WCT$^2$ with HFCS turned on (e). The proposed INSL (f) can overcome the side effect of SC while enjoying enhanced detail preservation.}
    \label{fig:3}
\end{figure}

\noindent\textbf{Feature Aggregation.} Feature aggregation is a network module that concatenates multi-scale features produced by different layers of deep networks. Feature aggregation enables networks to integrate information from different field-of-views, thus may enhance low-level detail preservation of stylization that happens in high-level features. Based on this, we introduce a bottleneck feature aggregation (BFA) module to the auto-encoder. In detail, we first resize features from ReLU\_1\_1 to ReLU\_4\_1 to the size of ReLU\_5\_1 in the VGG encoder, then we concatenate them together at the bottleneck. Please refer to Fig.~\ref{fig:2} (b) for details. We show the style transfer results produced by networks \emph{with} and \emph{without} BFA in Fig.~\ref{fig:2} (d) and (e) respectively, which show that BFA can preserve more fine details (\eg, more detailed tree branches and leaves in Fig.~\ref{fig:2}). To the best of our knowledge, we are the first that adopt the feature aggregation module to style transfer tasks.

\begin{figure}[t]
    \centering
    \includegraphics[width=\linewidth]{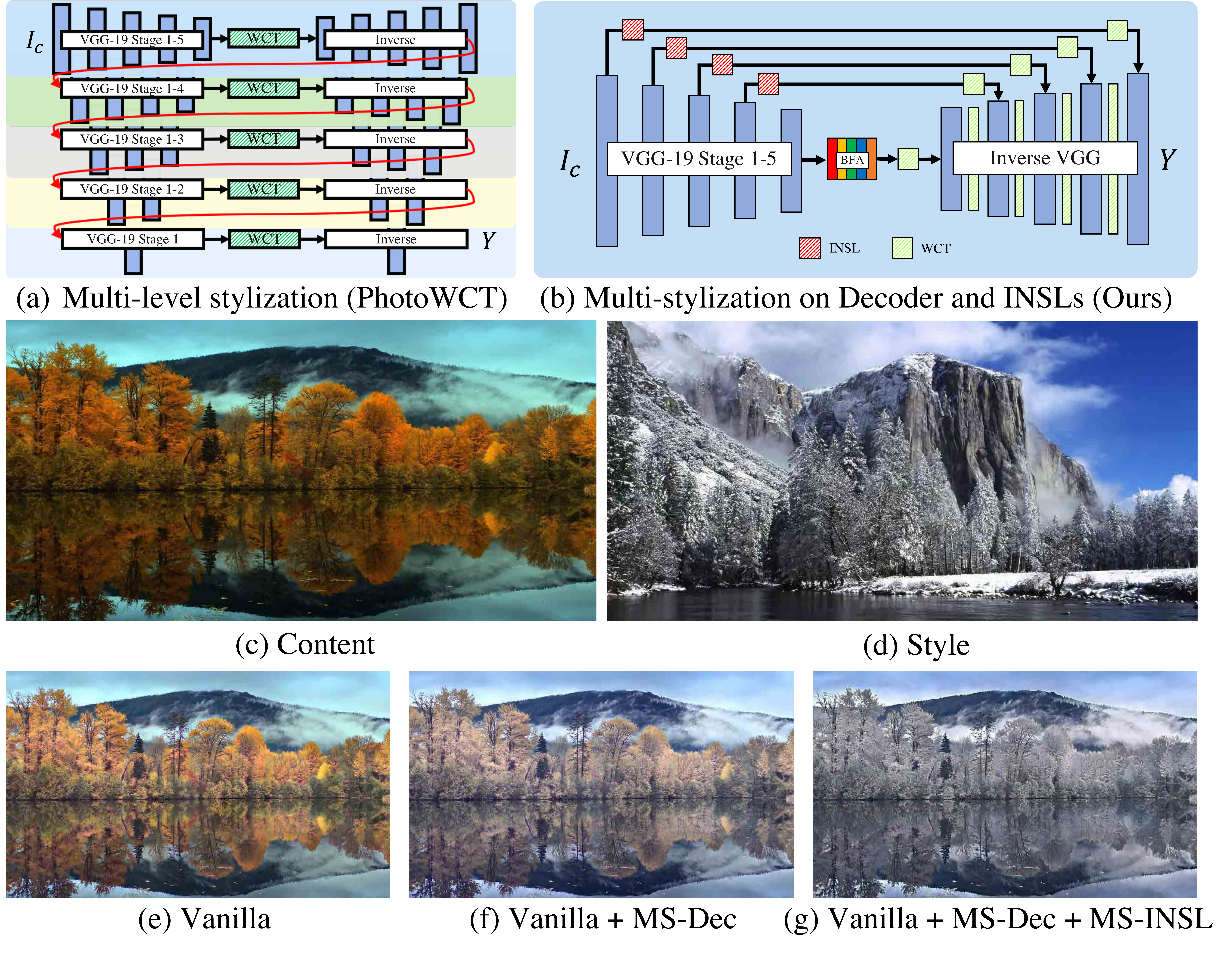}
    \caption{\textbf{Multi-stylization Comparison.} (a) is the multi-level stylization strategy used by WCT/PhotoWCT, which adopts five distinct auto-encoders in cascade to make style transfer. (b) is the architecture of our method. Please note that (b) equals to the auto-encoder in the top blue box in terms of computation cost. From (e) to (g), we progressively apply style transfer modules (\ie WCT) at the bottleneck, decoder, and INSLs, where MS-Dec and MS-INSL denote placing transfer module at decoder and INSLs respectively. As demonstrated in (e-g), MS-Dec and MS-INSL enhance style transfer effects without sacrificing fine details of the content. Please see colors of leaves in (e-g).}
    \label{fig:4}
\end{figure}
\begin{figure}[t]
    \centering
    \includegraphics[width=\linewidth]{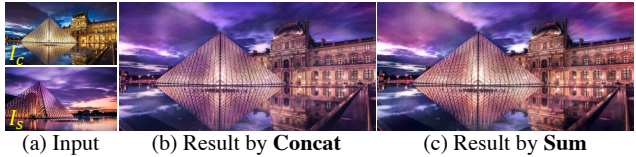}
    \caption{\textbf{Comparison of ``Concat'' and ``Sum''.}}
    \label{fig:5}
\end{figure}
\begin{figure}[t]
    \centering
    \includegraphics[width=\linewidth]{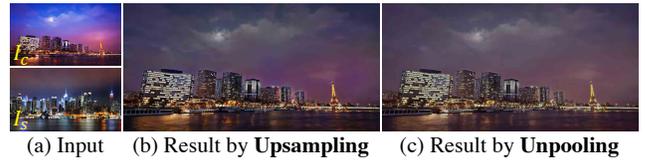}
    \caption{\textbf{Comparison of ``Upsampling'' and ``Unpooling''.}}
    \label{fig:6}
\end{figure}
\begin{figure}[t]
    \centering
    \includegraphics[width=\linewidth]{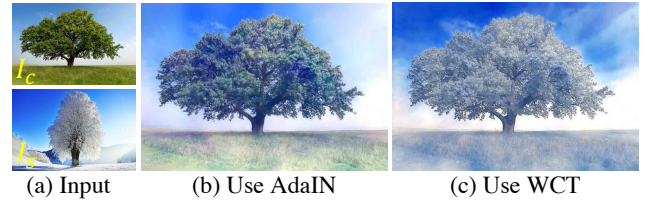}
    \caption{\textbf{Comparison of using AdaIN and WCT as transfer module.} Using WCT as transfer module (c) achieves more faithful photorealistic stylization effects against using AdaIN (b).}
    \label{fig:7}
\end{figure}
\begin{figure*}[t]
    \centering
    \includegraphics[width=0.75\textwidth]{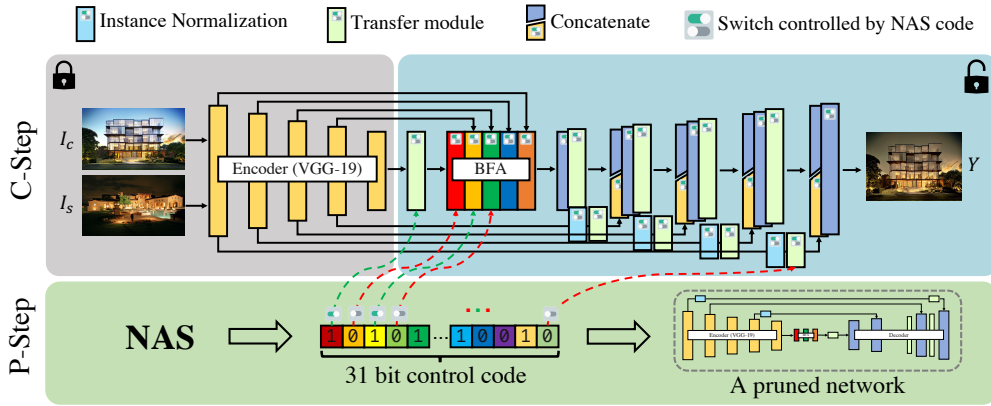}
    \caption{\textbf{Framework of the proposed method.} Our method consists of a C-step and P-step. In C-step, we construct a highly effective dense auto-encoder. In P-step, we propose a neural architecture search (StyleNAS) algorithm to automatically prune the auto-encoder. In each loop of the auto-pruning, the encoder part (in grep box) is fixed while 31 operators in the blue box is controlled by 0/1 code to turn off/on. Please note that yellow and cyan rectangles represent sequential convolution operators.}
    \label{fig:8}
\end{figure*}

\noindent\textbf{Skip Link.} The Skip Connection (SC) is first introduced by FCN~\cite{long2015fully} and U-Net~\cite{ronneberger2015u}, where SC can significantly enhance their segmentation results. However, the auto-encoder equipped with SC generally lost its ability to produce stylized images since SC can make the transfer module at the bottleneck of the auto-encoder invalid. We call this issue \emph{``short circuit''} phenomenon. As demonstrated in Fig.~\ref{fig:3} (d), the image produced by the auto-encoder with SC totally lost stylization effects compared with that without SC (show in Fig.~\ref{fig:3} (c)). The reason behind this is that SCs placed at low-level layers of an auto-encoder will short circuit and block the information stream flow into transfer modules work at the bottleneck. Interestingly, as shown in Fig.~\ref{fig:3} (e), we find that WCT$^2$ also fails to make stylization if turn their proposed High-Frequency Components Skip Links (HFCS) on and disable the input region masks. To solve this problem, we introduce the Instance Normalized Skip Links (namely INSL) as a replacement of the SC, which applies the Instance Normalization~\cite{ulyanov1607instance} at skip connections. We find that INSL can alleviate the short circuit phenomenon and strengthen the detail preservation and distortion elimination abilities of photorealistic style transfer networks. Please refer to Fig~\ref{fig:3} (f) for the result produced with INSLs.

\noindent\textbf{Multi-stylization.} Multi-stylization means make style transfer repeatedly. As shown in Fig.~\ref{fig:4} (a), WCT and PhotoWCT adopt a strategy called \emph{multi-level stylization}. They train five auto-encoders and make stylization for five rounds in a coarse-to-fine manner. Instead of that, WCT$^2$ proposes \emph{progressive stylization}, which uses a single round auto-encoder but progressively executes style transfer modules multi times at every part of the auto-encoder. Following WCT$^2$, we adopt a single-round multi stylization strategy but only transfer features at the decoder and INSLs. Fig.~\ref{fig:4} (b) illustrates our strategy. As demonstrated in Fig.~\ref{fig:4} (e-g), MS-Dec and MS-INSL can significantly improve the produced results in terms of stylization effects. Moreover, applying style transfer modules at INSLs (Fig.~\ref{fig:4} (g)) can further eliminate the short circuit phenomenon caused by SC and strengthen the stylization effects.

\noindent\textbf{Concat \vs Sum.} The choice of ``concat'' and ``sum'' operators when using skip links is a factor that may influence the performance of auto-encoders. However, we find that using ``concat'' generally has no specific difference against using ``sum'' except little style fluctuation. Please refer to Fig.~\ref{fig:5} (b) (c) for comparison.

\noindent\textbf{Upsampling \vs Unpooling.} PhotoWCT argues that the unpooling tends to make the network produce fewer distortions. However, we find that these two operators produce almost the same results in our settings. Please refer to Fig.~\ref{fig:6} (b) (c) for comparison.

\noindent\textbf{WCT \vs AdaIN.} WCT and AdaIN are two widely used transfer modules that come from artistic style transfer. As demonstrated in Fig.~\ref{fig:7} (b) (c), WCT can produces more faithful transfer results. We think this is because AdaIN need to work with the auto-encoder trained in a more complicated way. However, we just train the decoder to reconstruct images to facilitate the following pruning step.

\begin{table}[t]
    \caption{\textbf{Differences between our approach and other methods.}}
    \centering
    \tabcolsep=0.11cm
    \begin{tabular}{lcccc}
        \toprule
         & DPST & PhotoWCT & WCT$^2$ & Ours \\
        \midrule
        Learning-free & $\times$ & $\checkmark$ & $\checkmark$ & $\checkmark$\\
        No post-processing & $\checkmark$ & $\times$ & $\checkmark$ & $\checkmark$ \\
        No pre-mask & $\times$ & $\checkmark$ & $\times$ & $\checkmark$ \\
        Efficient & $\times$ & $\times$ & $\checkmark$ & $\checkmark$\\
\bottomrule
    \end{tabular}
\label{tab:2}
\end{table}
\section{C-Step}
Based on the analysis on architecture components that have significant influence on photorealistic style transfer effects, we construct an auto-encoder named \emph{PhotoNet}.

The C-step part (\ie, grey and blue boxes) in Fig.~\ref{fig:8} shows the architecture of PhotoNet. The encoder of PhotoNet is a VGG-19 that pre-trained on ImageNet dataset. The decoder is trained to invert deep features of the encoder back to images. In the bottleneck of PhotoNet, as demonstrated in the pre-analysis part, we place a BFA module to make use of multi-scale features. Between the encoder and decoder, we introduce INSLs to transport information from encoder stages (ReLU\_1\_1 to ReLU\_4\_1 in VGG-19) to their corresponding decoder layers. our INSL has two advantages: on the one hand, INSL enhances the detail preservation ability of PhotoNet, hereby improves photorealism. On the other hand, the equipped instance normalization can surprisingly weaken short circuit issue caused by skip connections. To improve photorealistic style transfer performance, we densely apply transfer modules (\ie, WCT) at the bottleneck, every stage of the decoder, and INSLs. Interestingly, making style transfer at INSLs further eliminated the short circuit phenomenon caused by skip links. 

During training, all transfer modules are temporarily removed and the encoder is fixed. The decoder (without transfer modules) is trained on MS\_COCO dataset~\cite{lin2014microsoft} to invert deep features of the encoder back to images. With the trained network, our PhotoNet directly takes a content photo and a style photo as input and outputs a style transferred photo. It is worth mentioning that our PhotoNet and the pruned version that will be introduced in the next part do not need any pre-conditioned region masks as DPST and WCT$^2$ do. Thanks to the strong detail preservation ability, our method enjoys fewer distortions against state-of-the-art algorithms while avoiding the usage of any time-consuming post-processing. Based on above-mentioned advantages, PhotoNet allows end-to-end photorealistic style transfer. 

Please refer to Fig.~\ref{fig:4} (g) for results of fully equipped PhotoNet. More results are available in supplementary materials. It is worth mentioning that PhotoNet is 7 and 107 times faster than WCT$^2$ (without counting the time for making segmentation masks) and PhotoWCT respectively. 
\begin{figure*}[t]
    \centering
    \includegraphics[width=0.85\textwidth]{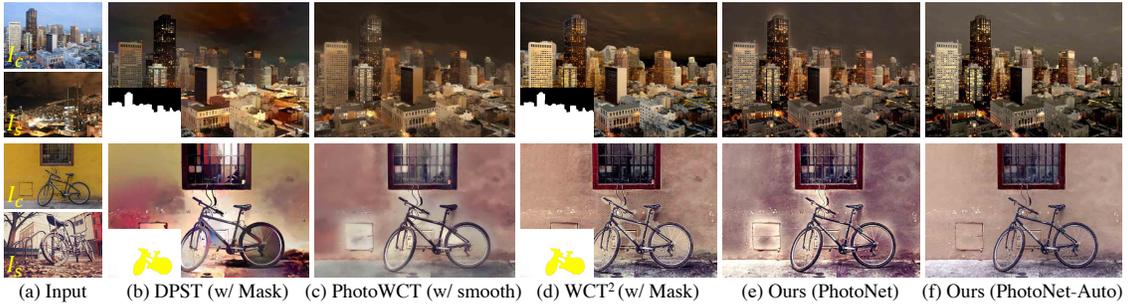}
    \caption{\textbf{Visual comparison to state-of-the-art methods.} (a) is the input content ($I_c$) and style ($I_s$) photos. DPST (b) and WCT$^2$ (d) have to run with the assist of regional masks (show in left-bottom corner) and the result of PhotoWCT (c) are produced with post-processing. Our methods ((e) PhotoNet, (f) PhotoNAS) do not need any pre- and post-processing.}
    \label{fig:9}
\end{figure*}
\begin{table*}[t]
    \caption{\textbf{Quantitative evaluation results for stylization methods.} Higher SSIM-Edge and SSIM-Whole scores mean the measured image is more similar to the input content photo in terms of fine details. A lower Gram Loss denotes the evaluated image has more similar visual effects to the style photo. Here results of DPST and WCT$^2$ are produces without the assist of segmentation maps for a fair comparison.}
\centering
    \begin{tabular}{lcccccc}
        \toprule
        Method & DPST & PhotoWCT & PhotoWCT+Smooth & WCT$^2$ & Ours(PhotoNet) & Ours(PhotoNAS) \\
        \midrule
        SSIM-Edge~$\uparrow$ & 0.6395 & 0.5690 & 0.6391 & 0.6112 & 0.6922 & \textbf{0.6932} \\
        SSIM-Whole~$\uparrow$ & 0.5139 & 0.5013 & 0.5005 & 0.4723 & \textbf{0.7047} & 0.6728\\
        Gram Loss~$\downarrow$ & 1.4143 & 1.2130 & 2.1660 & \textbf{1.1244} & 1.1270 & 1.7565 \\
        \bottomrule
    \end{tabular}
    \label{tab:1}
\end{table*}

\section{P-Step}
To further accelerate PhotoNet, a P-step is proposed to automatically prune PhotoNet and discover more efficient style transfer networks for photorealistic rendering while maintaining stylization effects of the PhotoNet. We achieve this by using PhotoNet as the maximum architecture and introducing a neural architecture search method named StyleNAS in a manner of teacher-student learning for automatic pruning. Given the MS\_COCO as the training dataset and a validation dataset with 40 content and style photos, we first train PhotoNet as the \emph{Supervisory Oracle} for the subsequent architecture search. The P-step consists of the following three key components. 

\noindent\textbf{Search Space.} We use fully equipped PhotoNet and all of its simplified versions (remove some operators) as the search space. Please refer to grey and blue boxes in Fig.~\ref{fig:8} (\ie C-step part) for details. In each loop of the neural architecture search, 31 options of operators have been remained to form a functional architecture, while one can open/close a bit to determine use/ban an operator. We encode any architecture in this space using a string of $31$-bits. For example, the searched PhotoNAS architecture is encoded as ``0101000000100000000000000001111'' in our setting. In this way, StyleNAS can search new architectures in a combinatorial manner from totally $2^{31}\approx 2.1\times 10^9$ possible architectures. We hereby denote the search space as $\Theta$ which refers to the full set of all architectures. 

\noindent\textbf{Search Objectives.} To obtain highly-efficient and effective architectures from $\Theta$, we adopt three search objectives: (i) the loss of knowledge distillation from a pre-trained supervisory oracle (PhotoNet), (ii) the perceptual loss of the produced images and oracle, and (iii) the percentage of operators used in the architecture.  The knowledge distillation loss reflects image reconstruction errors in a supervisory manner. We write the overall search objective as

\begin{align}
    \mathcal{L}(\theta) &= \alpha\cdot\mathcal{E}(\theta) + \beta\cdot\mathcal{P}(\theta)+\gamma\cdot\mathcal{O}(\theta),\\
    \mathcal{E}(\theta) &= \underset{I \in \mathbb{V}}{\mathrm{mean}}\ \|I_{\theta} - I_{oracle}\|_F,\\
    \mathcal{P}(\theta) &= \underset{I \in \mathbb{V}}{\mathrm{mean}}\ \sum\limits_{i=1}^5 \|\Phi_i\left( I_{\theta} \right) - \Phi_i\left( I_{oracle} \right)\|_F,
    \label{eq:obj}
\end{align}

where $\theta\in\Theta$ refers to an architecture drawn from the space; $\mathcal{L}(\theta)$ stands for the overall loss of the architecture $\theta$; $\mathcal{E}(\theta)$ refers to the reconstruction error between the style-transferred images produced by the network with the architecture $\theta$ and those produced by the supervisory oracle; $\mathcal{P}(\theta)$ estimates the \emph{Perceptual Loss} using a trained network with the architecture $\theta$ and the oracle; $\Phi_i\left( \cdot \right)$ denotes the output of the $i^{th}$ stage of the ImageNet pre-trained VGG-19; $\mathbb{V}$ denotes the validation set with 40 content and style photos; $\mathcal{O}(\theta)$ estimates the percentage of operators used in $\theta$ of $31$-bins; $\alpha, \beta$ and $\gamma$ are a pair of hyper-parameters to make trade-off between these three factors. 

\noindent\textbf{Search Strategies.} Our search strategies are derived from~\cite{kim2017nemo}, where parallel evolutionary strategies with a map-reduce alike update mechanism have been used to iteratively improve the searched architectures from random initialization. From the search space $\Theta$, the StyleNAS algorithm first randomly draws $P$ architectures $\{\theta^1_1,\theta^2_1,\theta^3_1\dots\theta^P_1\}\subset\Theta$ (represented as P $31$-bit strings) for the $1^{st}$ round of iteration, where $P$ refers to the number of populations desired. On top of the parallel computing environment, the algorithm maps every drawn architecture to one specific GPU card/worker, then trains the style transfer networks for image reconstruction (with WCT modules temporarily turned off), and evaluates the performance of trained networks (using the objectives in Eq~\ref{eq:obj}). With the search objective estimated, every worker updates a shared \emph{population set} using the evaluated architecture in an asynchronous manner, and generates a new architecture through \emph{mutating} the best one in a subset of architectures drawn from the \emph{population set}. With the newly generated architecture, the worker starts a new iteration of training and evaluating for the update and discards the oldest model from the \emph{population set}. During the whole process, the algorithm keeps maintaining a \emph{history set} of architectures that have been explored with their objectives estimated, all in an asynchronous manner. After $T$ rounds of iterations on every worker, the algorithm returns the architecture with the minimal objectives from the overall \emph{history set} by the end of the algorithm. Please refer to the supplementary for more details.

\section{Experimental Results}
In this section, we show the result comparison of our algorithm with state-of-the-art photorealistic stylization methods, i.e., DPST, PhotoWCT, and WCT$^2$ in terms of visual effects and time consumption. More comparison results, detailed experimental settings, user study results, video transfer results, and our failure cases are available in supplementary materials. All the source code will be made released in the future.

\noindent\textbf{Visual Comparison}
We testify the effectiveness of the proposed method by the comparison with the photorealistic stylization results of DPST, PhotoWCT, and WCT$^2$. Since the official code of DPST and WCT$^2$ require pre-acquired regional masks as the input arguments, we make comparison on images and corresponding segmentation masks provided by DPST in this part. Additionally, we add two post-processing steps to PhotoWCT as suggested by its paper. Note that results of our approaches (PhotoNet and PhotoNAS) do not involve any pre- and post-processing.

As shown in Fig.~\ref{fig:9}, results of DPST contains significant artifacts and are comparably over-smoothed. For example, textures of buildings in the upper photo and details of bicycle wheels in the bottom image are blurred. Moreover, wall and ground in the bottom image show undesirable colors. Although results of PhotoWCT (w/ smooth) (Fig.~\ref{fig:9} (c)) have alleviated artifacts, they still suffer from distortions and create blurry images since they have to use smooth-oriented post-processing to decrease those artifacts. WCT$^2$ make some advances upon previous two methods in terms of detail preservation by applying regional masks. However, WCT$^2$ introduces a new drawback that the produced images usually have visual style mismatch at the boundary of different regions. Even worse, if those masks are not accurate enough, WCT$^2$ tends to generate images with considerable artifacts which significantly hurt the photorealism of produced images. Please \emph{zoom-in} in Fig.~\ref{fig:9} (d) to see skylines in the upper example and bicycle outlines painted on the wall in the bottom result. Foregrounds of the result by WCT$^2$ look like are pasted on the background, which is non-photorealistic. Fig.~\ref{fig:9} (e) and (f) show results of our methods. PhotoNet achieves effective photorealistic stylization and faithful detail preservation. The results of PhotoNAS maintains the stylization effects of PhotoNet and in the meantime, further eliminates remained distortions. It is worth mentioning that PhotoNAS achieves such a result with only 1/5 time-consumption. Note that results of PhotoNet and PhotoNAS are produced without any pre- and post-processing while other methods use pre- (DPST, WCT$^2$) or post-processing (PhotoWCT). Please refer to Fig.~\ref{fig:1} for comparison without pre-/post-processing, which additionally verified the effectiveness of our method.

\noindent\textbf{Quantitative Comparison.}
Inspired by WCT$^2$, we adopt structural similarity (SSIM) index between the original content photo and the produced result to measure the detail preservation ability (\ie photorealism) of methods. We compute SSIM on whole images (named SSIM-Whole) and their holistically-nested edge responses~\cite{xie2015holistically} (named SSIM-Edge). To evaluate photorealistic stylization effects, we compute the Gram matrix difference (VGG style loss) following WCT.

Given a validation dataset contains 73 content and style photo pairs, we quantitatively evaluate the performance of the proposed and state-of-the-art methods by computing the above-mentioned metrics on this validation set. We show the quantitative comparison result in Tab.~\ref{tab:1}. The proposed PhotoNet and PhotoNAS achieve better scores in terms of SSIM-Whole and SSIM-Edge respectively, which means our methods have remarkably improved detail preservation ability. Tab~\ref{tab:1} shows that the Gram Loss of our PhotoNet is a little higher than WCT$^2$. We argue this is due to the improvement of detail preservation would inevitably raise the Gram Loss. Such an assertion is also verified by the fact that the Gram Loss of PhotoWCT largely increased when applying smooth post-processing.

\begin{table}[t]
    \caption{\textbf{Computing-time comparison.}}
            \vspace{-3mm}
    \centering
    \small
    \tabcolsep=0.08cm
    \begin{tabular}{lccccc}
        \toprule
        Method & DPST & PhotoWCT & WCT$^2$ & PhotoNet & PhotoNAS \\
        \midrule
        $256\times128$ & 114.11 & 4.07 & 4.42 & 0.76 & \textbf{0.13} \\
        $512\times256$ & 293.28 & 20.72 & 5.28 & 0.86 & \textbf{0.16} \\
        $768\times384$ & 628.24 & 53.05 & 6.30 & 0.95 & \textbf{0.22} \\
        $1024\times512$ & 947.61 & 133.90 & 7.69 & 1.06 & \textbf{0.32} \\
\bottomrule
    \end{tabular}
        \vspace{-3mm}
\label{tab:3}
\end{table}

\noindent\textbf{Computational Time Comparison.}
We conduct a computing time comparison against the state-of-the-art methods to demonstrate the efficiency of the proposed and searched network architectures. All approaches are tested on the same computing platform which includes an NVIDIA P100 GPU card with 16GB RAM. The time consumption of DPST, PhotoWCT, and WCT$^2$ are evaluated by running officially released code with their default settings. We compare the computing time on content and style images with different resolutions. As Table~\ref{tab:2} shows, PhotoNet achieves $6\times$ faster against WCT$^2$ and PhotoNAS are almost 20-30$\times$ faster than WCT$^2$. Surprisingly, after the P-step, only 7 operators are left among searched ones.

\section{Conclusion}
In this paper, we present a two-stage method to address the photorealistic style transfer problem. In the first step, we analyze the influence of commonly used network architectural components on photorealistic style transfer. Based on that, we construct PhotoNet, which utilizes instance normalized skip links (INSL), bottleneck feature aggregation (BFA), and multi-stylization on decoder and INSLs, to generate rich-detailed and well-stylized images. In the P-step, we introduce a network pruning framework for photorealistic stylization adopting a neural architecture search (StyleNAS) method and teacher-student learning strategy. With the novel pruning method, we discover PhotoNAS, which is surprisingly simple and keeps the stylization effects almost intact. Our extensive experiments in terms of visual, quantitative, and computing time comparison show that the proposed approach has a strengthened ability to remarkably improve the stylization effects and photorealism while reducing the time consumption dramatically. Our study also expands the application area of NAS to photorealistic style transfer. In our future work, we plan to 1) design novel NAS method specifically for style transfer task and 2) extend the work to other generative models such as generative adversarial networks and other low-level vision tasks.

\begin{figure*}[t]
    \centering
    \includegraphics[width=\textwidth]{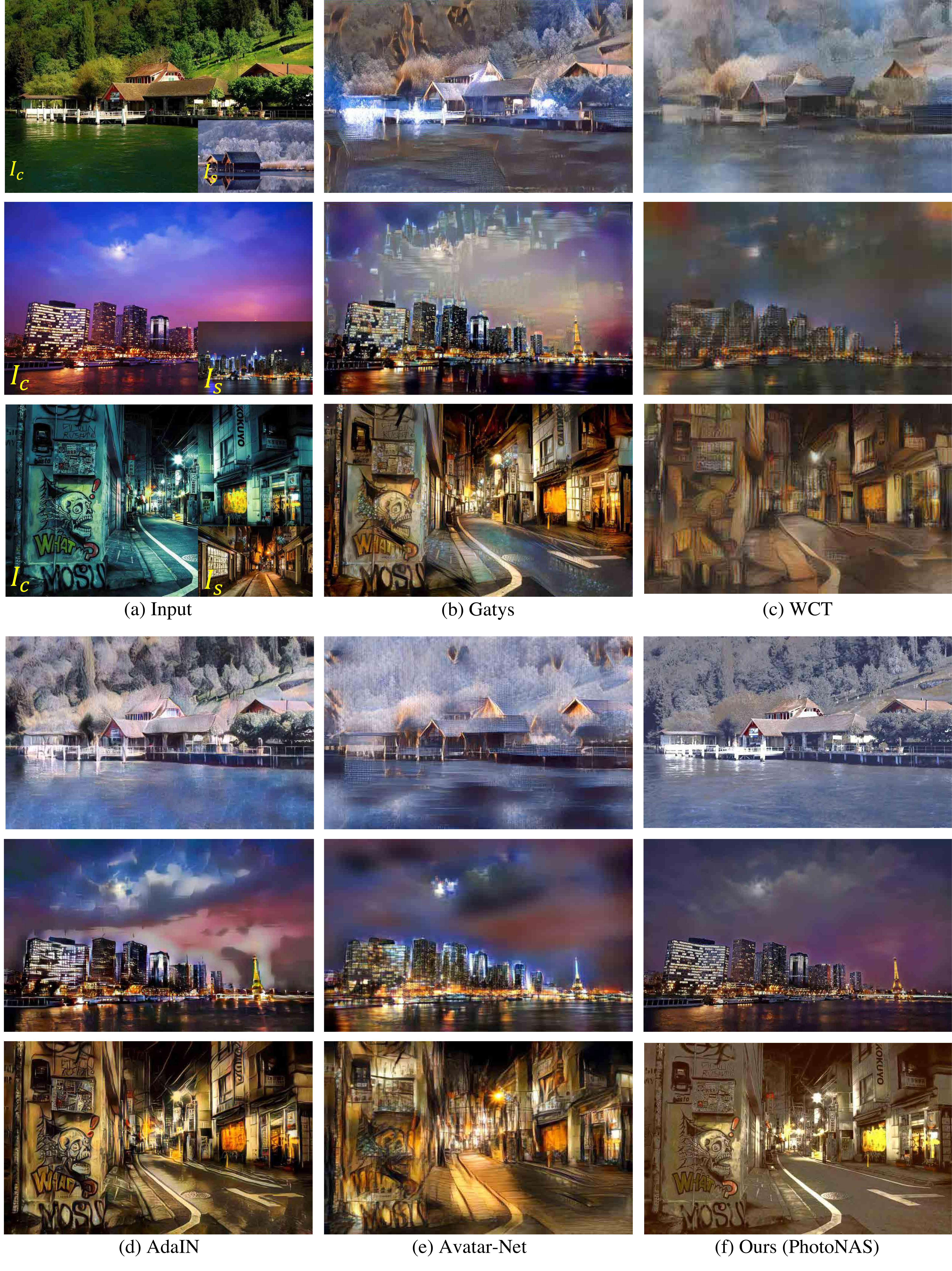}
    \caption{\textbf{Result of photorealistic style transfer by artistic style transfer methods.}}
    \label{appendixfig:1}
\end{figure*}

\section{Appendix A: More Photorealistic Stylization Results by Artistic Style Transfer Methods}
In this part, we show more photorealistic stylization results by using artistic style transfer approaches. Given 73 content and style photo pairs, we conduct experiment on Gatys~\cite{Gatys2016}, WCT~\cite{li2017universal}, AdaIN~\cite{huang2017arbitrary}, and Avatar-Net~\cite{sheng2018avatar}. As stated in our paper, artistic style transfer methods fail to make decent photorealistic stylization. We think the reason behind this is that artistic style transfer methods tend to pursue a comparably low gram loss (representing \textbf{artistic} style similarity) between the style photo and the produced image, which generally can enhance artistic style transfer effects in terms of the similarity of shapes, lines, colors, and textures. However, those inevitable changes to content details are not favored in photorealistic style transfer scenarios. Thus artistic style transfer methods usually fail to produce photorealistic photos. Fig~\ref{appendixfig:1} (b-e) show stylization results with state-of-the-art artistic style transfer methods: Gatys, WCT, AdaIN, and Avatar-Net. The produced images have significant artifacts and distortions in the perspective of photorealistic stylization. On the contrary, our method (PhotoNAS) specifically designed for photorealistic style transfer produces more photorealistic stylization result as demonstrated in Fig.~\ref{appendixfig:1} (f).

\begin{figure*}[t]
    \centering
    \includegraphics[width=1.0\textwidth]{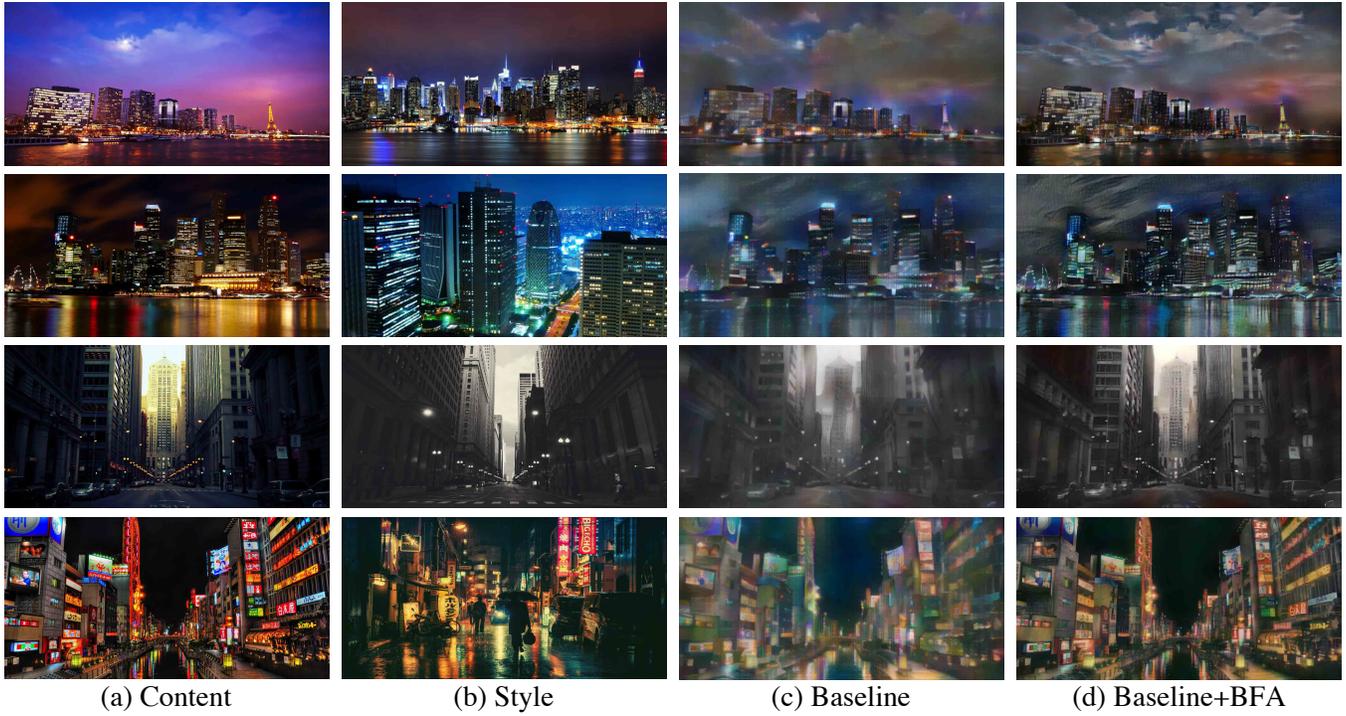}
    \caption{\textbf{Result comparison about Bottleneck Feature Aggregation (BFA).}}
    \label{appendixfig:2}
\end{figure*}
\begin{figure*}[t]
    \centering
    \includegraphics[width=1.0\textwidth]{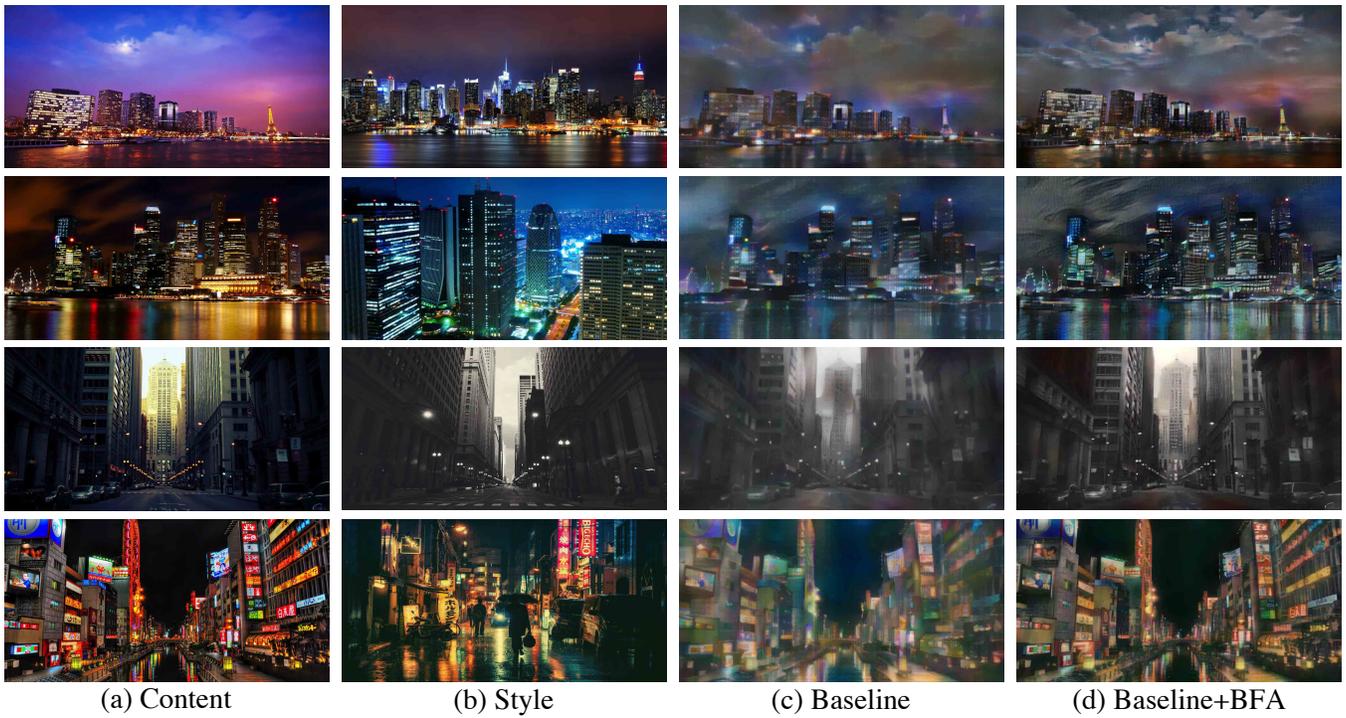}
    \caption{\textbf{Result comparison about Instance Normalized Skip Link (INSL).}}
    \label{appendixfig:3}
\end{figure*}
\begin{figure*}[t]
    \centering
    \includegraphics[width=\textwidth]{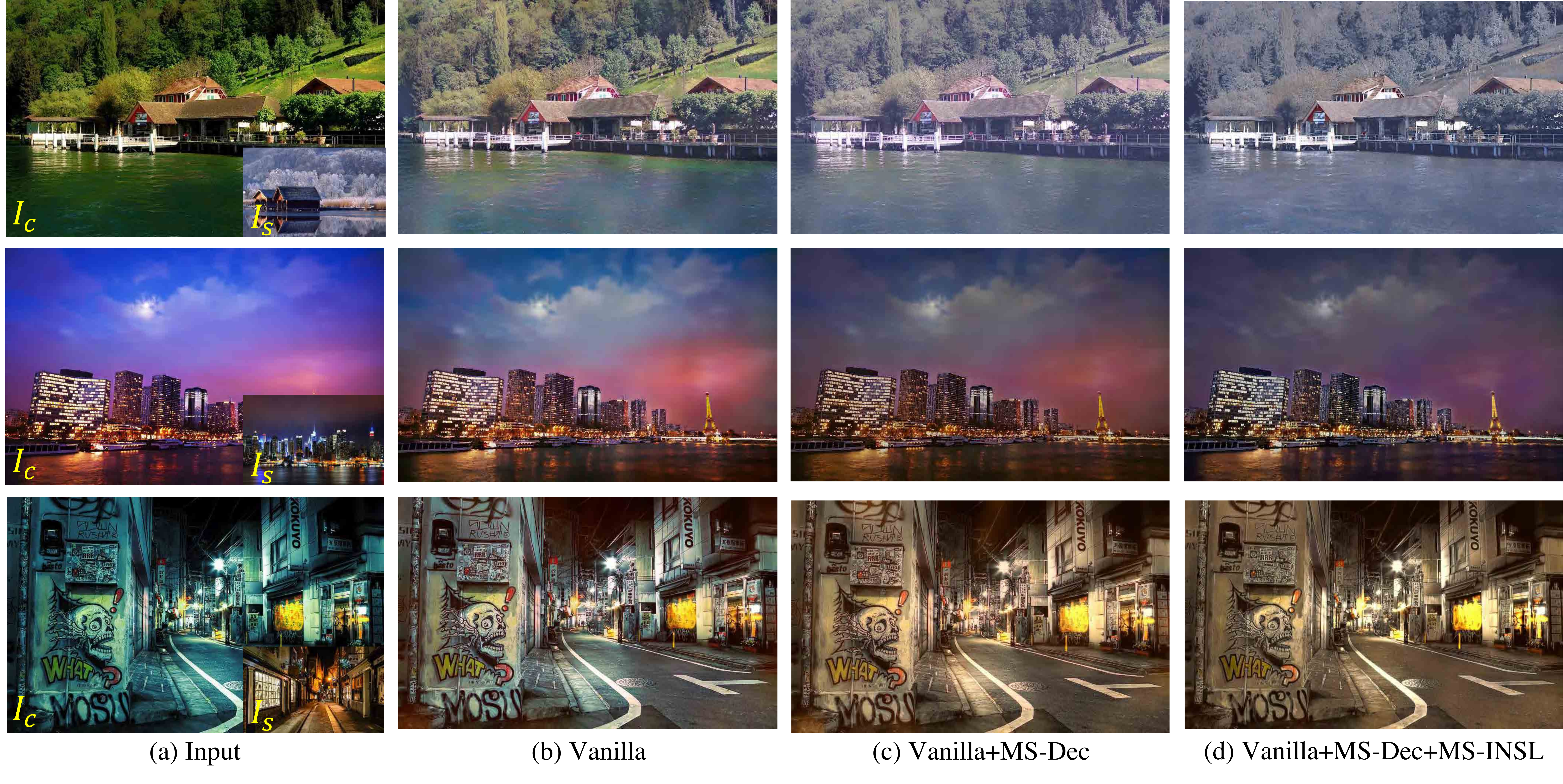}
    \caption{\textbf{Result comparison about Multi-Stylization strategy.}}
    \label{appendixfig:4}
\end{figure*}
\begin{figure*}[t]
    \centering
    \includegraphics[width=1.0\textwidth]{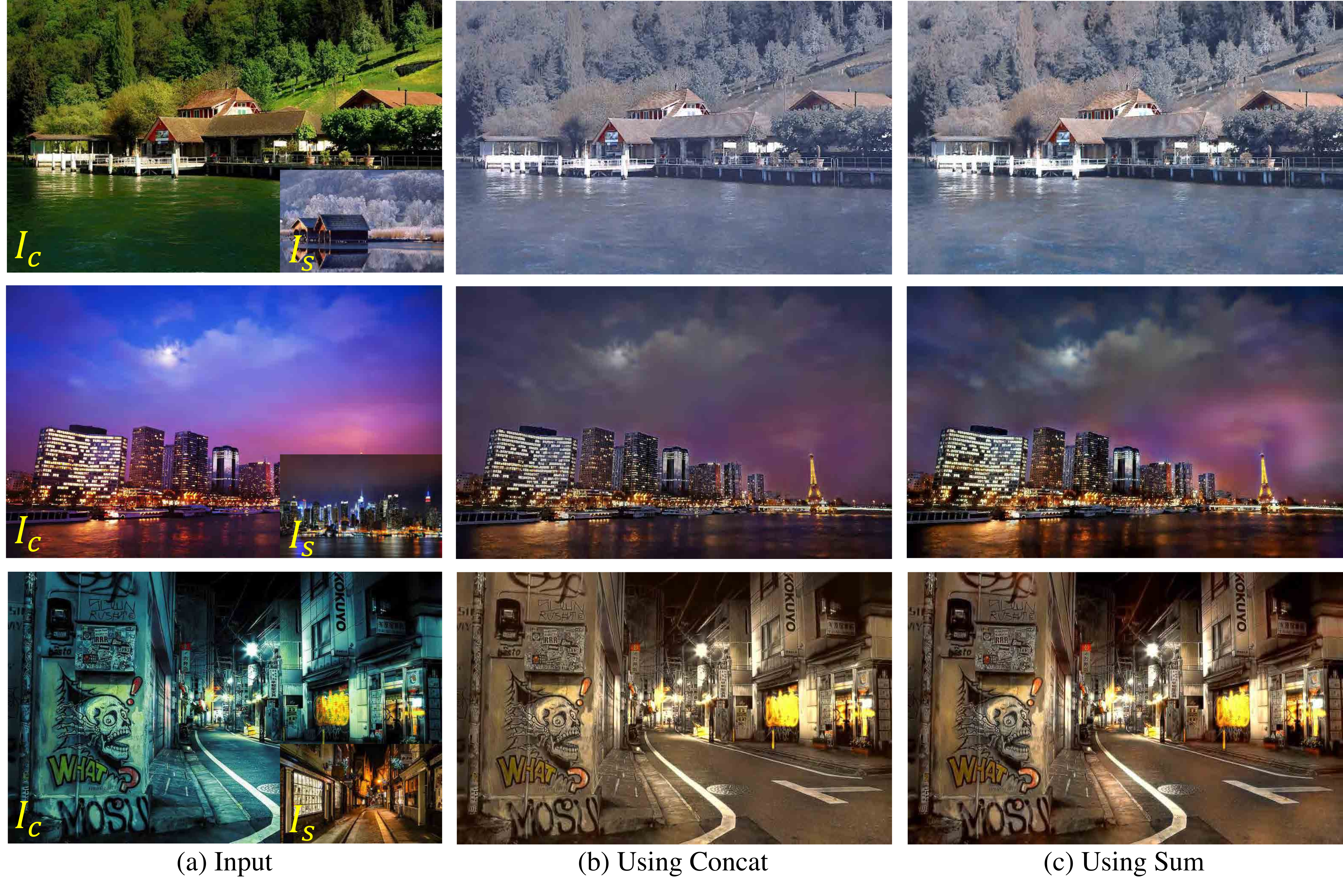}
    \caption{\textbf{Result comparison between using Concat and Sum.}}
    \label{appendixfig:5}
\end{figure*}
\begin{figure*}[t]
    \centering
    \includegraphics[width=\textwidth]{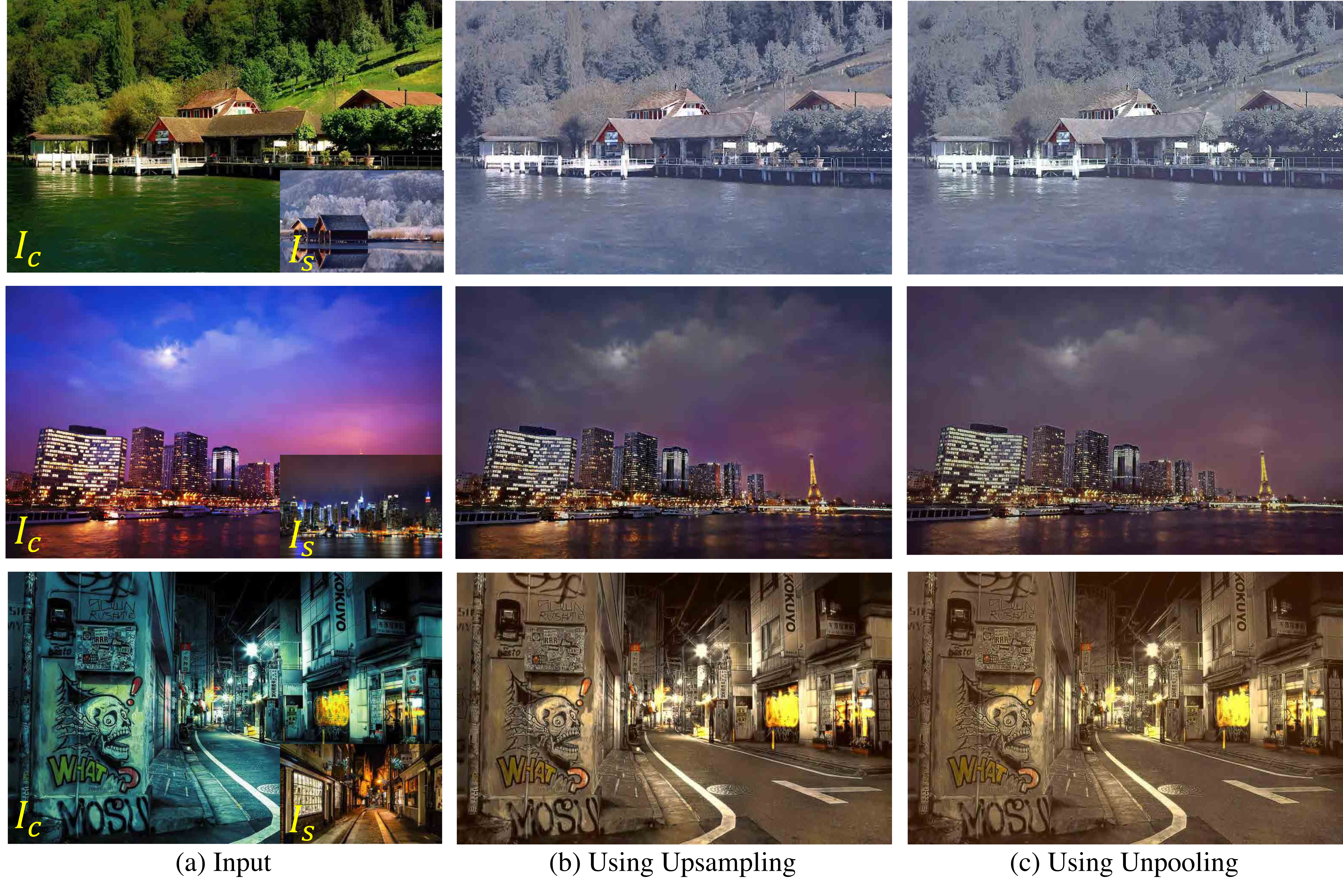}
    \caption{\textbf{Result comparison between using Upsampling and Unpooling.}}
    \label{appendixfig:6}
\end{figure*}
\begin{figure*}[t]
    \centering
    \includegraphics[width=1.0\textwidth]{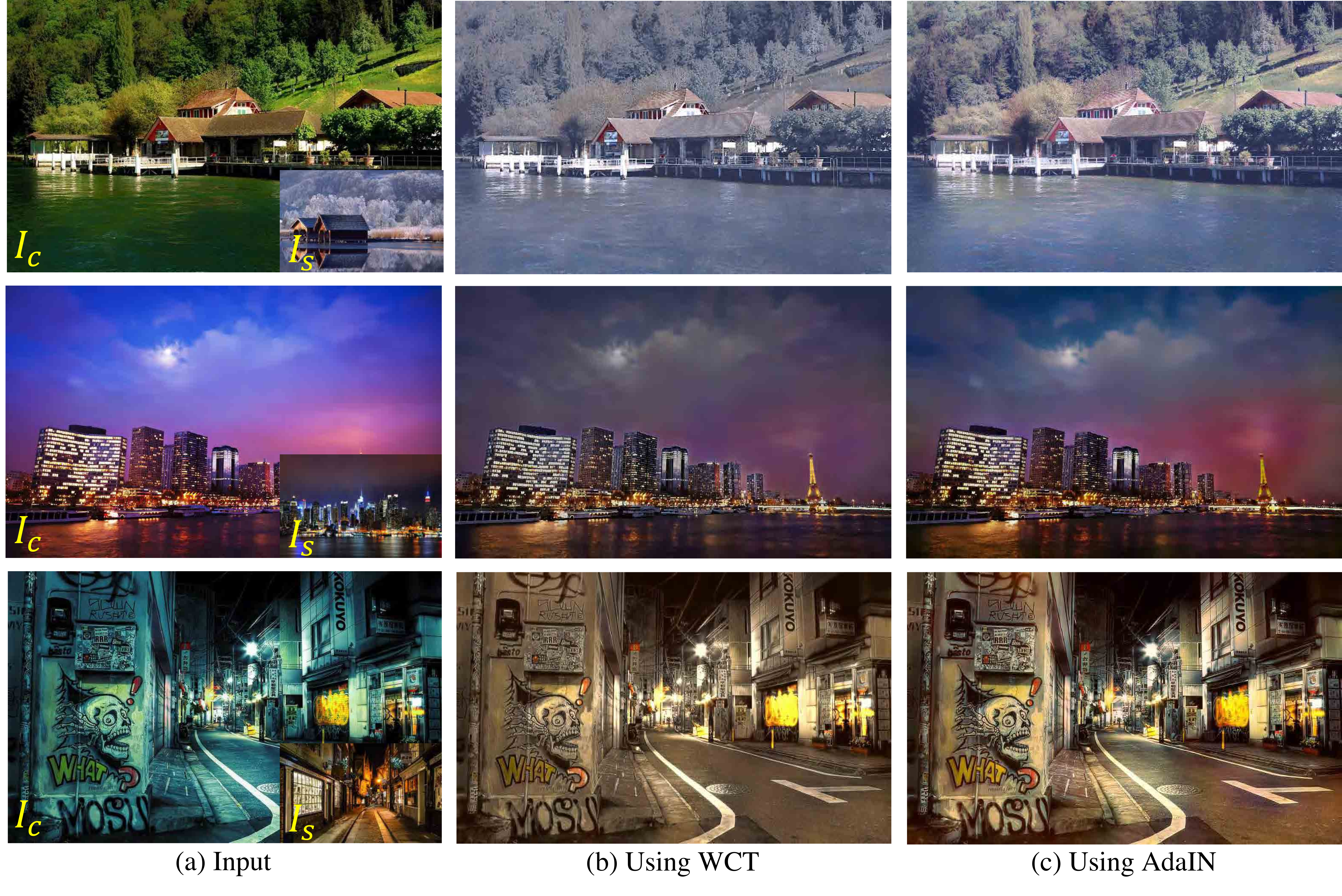}
    \caption{\textbf{Result comparison between using WCT and AdaIN.}}
    \label{appendixfig:7}
\end{figure*}
\section{Appendix B: More Results in Pre-Analysis}
In this part, we show more results of pre-analysis studying effective architectural modules for photorealistic style transfer.
\subsection{Feature Aggregation}
We are the first that introduce Bottleneck Feature Aggregation module to style transfer tasks. As demonstrated by Fig.~\ref{appendixfig:2}, BFA can remarkably improve photorealistic stylization effects.
\subsection{Skip Link}
As stated in our paper, the proposed INSL can address the ``short circuit'' issue caused by Skip Connections and HFCS in WCT$^2$. Fig.~\ref{appendixfig:3} shows more result comparison of Baseline (c), Baseline+SC (d), WCT$^2$+HFCS (e), and Baseline+INSL (f), which demonstrates the effectiveness of the proposed INSL.
\subsection{Multi-Stylization}
Fig.~\ref{appendixfig:4} demonstrates that MS-Dec and MS-INSL have the ability to enhance photorealistic stylization effects while maintain the content details intact.
\subsection{Concat\vs Sum}
Please refer to Fig.~\ref{appendixfig:5}.
\subsection{Upsampling \vs Unpooling}
Please refer to Fig.~\ref{appendixfig:6}.
\subsection{WCT\vs AdaIN}
Please refer to Fig.~\ref{appendixfig:7}.

\begin{algorithm}[t]
\caption{StyleNAS algorithm}
\small
\begin{algorithmic}
\State Train the \emph{supervisory oracle} ($\mathcal{SO}$) network;
\State Set overall search space $\Theta$, revolution cycle $C$, the population size $P$, the population/history sets $\Theta^{pop},\Theta^{history} \leftarrow \emptyset$, generation index $gen\gets 0$;
\While{$|\Theta^{pop}| < P$\ \textbf{in parallel} }
    \State $\theta\leftarrow$ \textproc{RandomArchitecture}($\Theta$);
    \State $\theta.loss\leftarrow\mathcal{L}(\theta)$ through training and evaluating a network based on $\theta$ and $\mathcal{SO}$; 
    \State $\theta.gen\leftarrow gen$;
    \State $\Theta^{pop}\leftarrow\Theta^{pop}\cup\{\theta\}$;
    \State$\Theta^{history}\leftarrow\Theta^{history}\cup\{\theta\}$;
\EndWhile
\While{$|\Theta^{history}|<C$}
\State $gen\gets gen+1$; 
    \For{$i < P$\ \textbf{in parallel}}
    \State Randomly pickup a \emph{subset} of architectures from $\Theta^{pop}$ as $\Delta^{pop}\subseteq\Theta^{pop}$;
    \State Set $\theta^{parent}\gets \underset{\theta\in\Delta^{pop}}{\mathrm{argmin}}\ \theta.loss$ using the architecture in $\Delta^{pop}$ with minimal loss;
    \State $\theta^{child}\leftarrow$ \textproc{Mutate}($\theta^{parent}$)
     \State $\theta^{child}.gen\leftarrow gen$;
     \State $\theta^{child}.loss\leftarrow\mathcal{L}(\theta^{child})$ through training and evaluating a network based on $\theta^{child}$ and $\mathcal{SO}$;
    \State $\Theta^{pop}\leftarrow\Theta^{pop}\cup\{\theta^{child}\}$;
    \State $\Theta^{history}\leftarrow\Theta^{history}\cup\{\theta^{child}\}$;
        \State Set $\theta^{oldest}\leftarrow \underset{\theta\in\Theta^{pop}}{\mathrm{argmin}}\ \theta.gen$ using the architecture in $\Theta^{pop}$ with minimal generation index;
    \State $\Theta^{pop}\leftarrow\Theta^{pop}\backslash\{\theta^{oldest}\}$;
    \EndFor
\EndWhile
\State \Return $\theta^{best}\leftarrow\underset{\theta\in\Theta^{history}}{\mathrm{argmin}}\ \theta.loss$ using the architecture in $\Theta^{history}$ with minimal loss;
\end{algorithmic}   
\label{alg:1}
\end{algorithm}

\begin{figure*}[t]
    \centering
\includegraphics[width=1.0\textwidth]{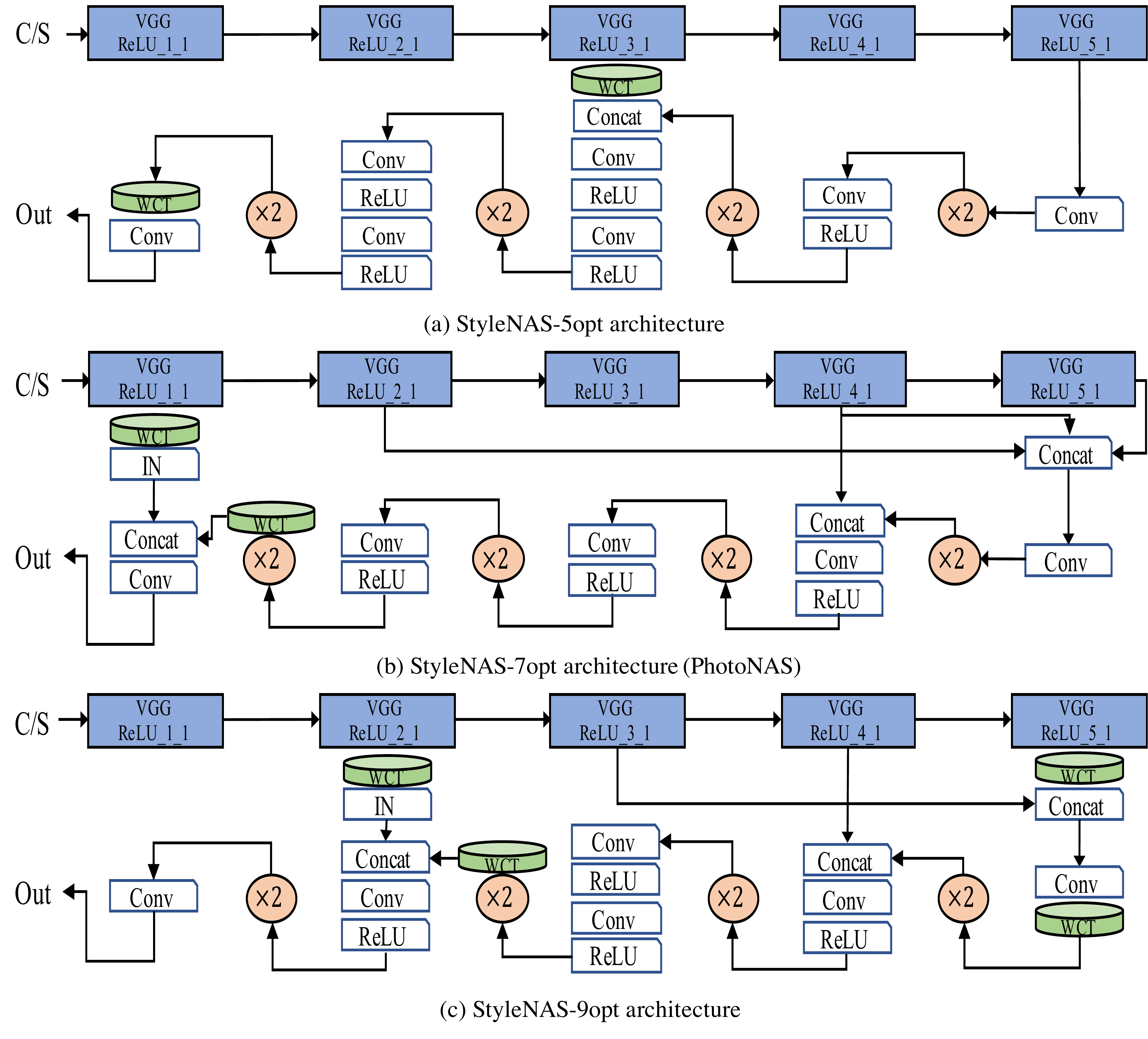}
\caption{\textbf{Searched architectures: StyleNAS-5opt, StyleNAS-7opt, and StyleNAS-9opt.}}
    \label{appendixfig:architecture}
\end{figure*}

\section{Appendix C: Details of P-Step}
\subsection{Implementation Details of StyleNAS Algorithm}
To search effective yet efficient network architectures for photorealistic stylization, we adopt an evolutionary algorithm to conduct a black-box optimization. Alg.~\ref{alg:1} shows our search method. We use up to 50 GPU cards during neural architecture search and each new architecture is trained and validated on a single GPU card. Our search method is based on \emph{Nemo} algorithm by~\cite{kim2017nemo}. We find this search algorithm is adequate to discover satisfactory network architectures in our framework. However, the search process by using original Nemo costs around 96 hours in searching for around new 190 architectures. To accelerate P-step, we modify the Nemo algorithm by changing the way to pick up new architectures from cascade to parallel. This parallel strategy works at the initialization period and the mutation of new architectures in every loop. Please see more details in the pseudo-code, where our modifications are marked in bold. Since the improved Nemo can make use of multi GPU cards in parallel for new architecture training and validation, it reduces 2/3 time-consumption of Nemo and cost around 24 hours to explore the same number of architectures. Theoretically, such a modification may fade search results. However, we find our modified Nemo successfully find one of the best architectures among search space while enjoying significant time-consumption reduction. Due to the huge computation cost and time-consumption (even after acceleration) of using NAS, we do not try other search methods. Moreover, this method already achieves satisfactory results. However, we will try other search strategies in our future work. 

\subsection{Experimental Settings of StyleNAS}
In the experiments that we reported, the hyper-parameters of StyleNAS are set as $\alpha=0.8$, $\beta=0.1$, and $\gamma=0.1$ to search highly efficient and effective architectures. Through the linear combination of three objective functions introduced in our paper, the StyleNAS algorithm found a group of time-efficient network architectures without stylization effects compromised. 

\begin{figure*}
    \centering
    \subfloat[Content]{\includegraphics[width=0.24\linewidth]{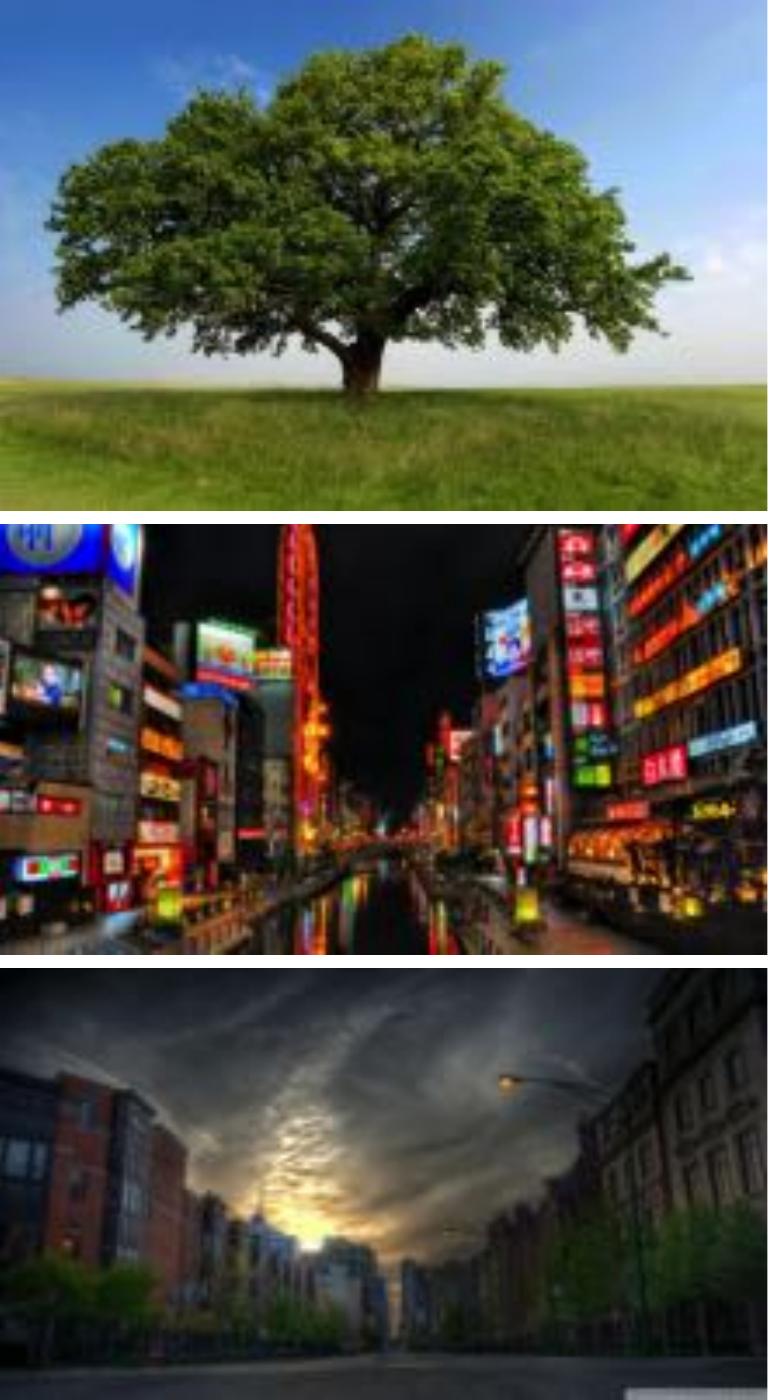}}\ 
    \subfloat[Style]{\includegraphics[width=0.24\linewidth]{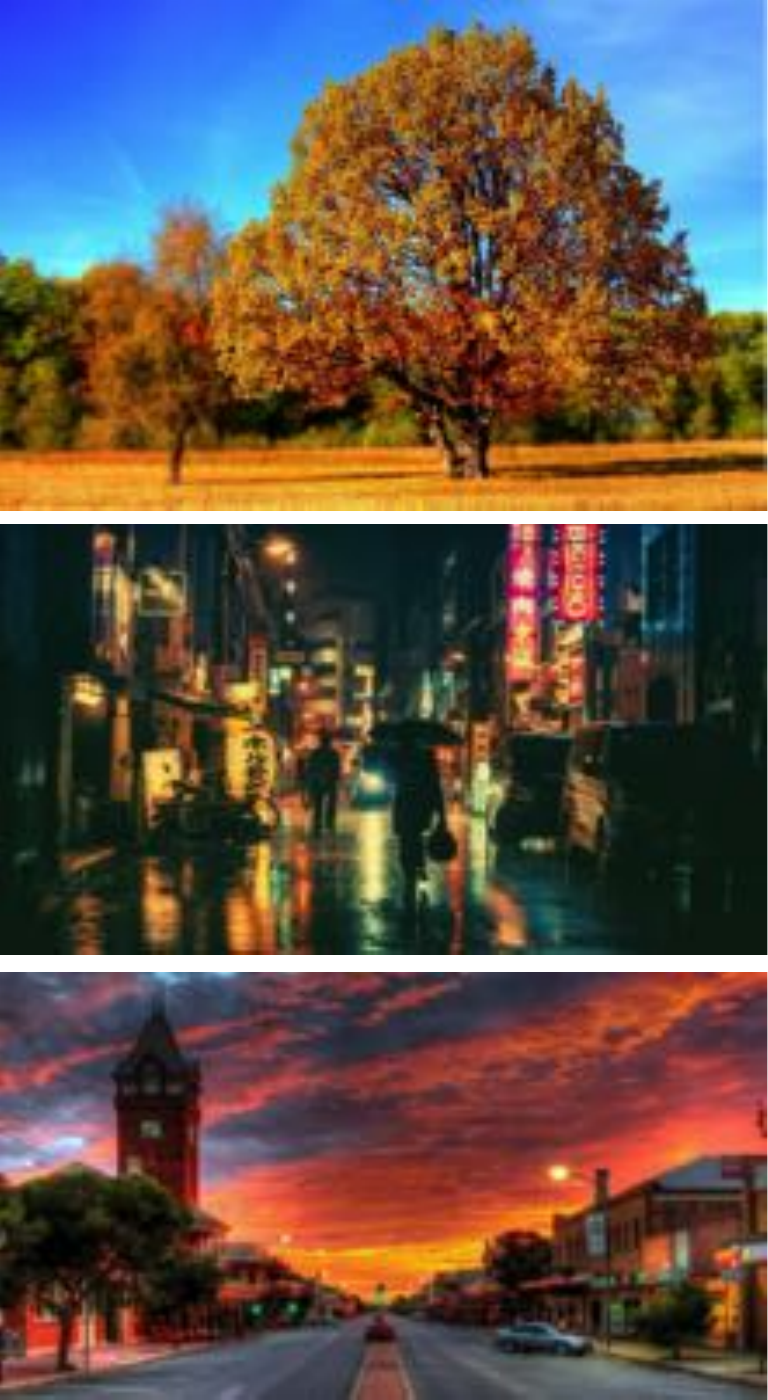}}\ 
    \subfloat[PhotoWCT]{\includegraphics[width=0.24\linewidth]{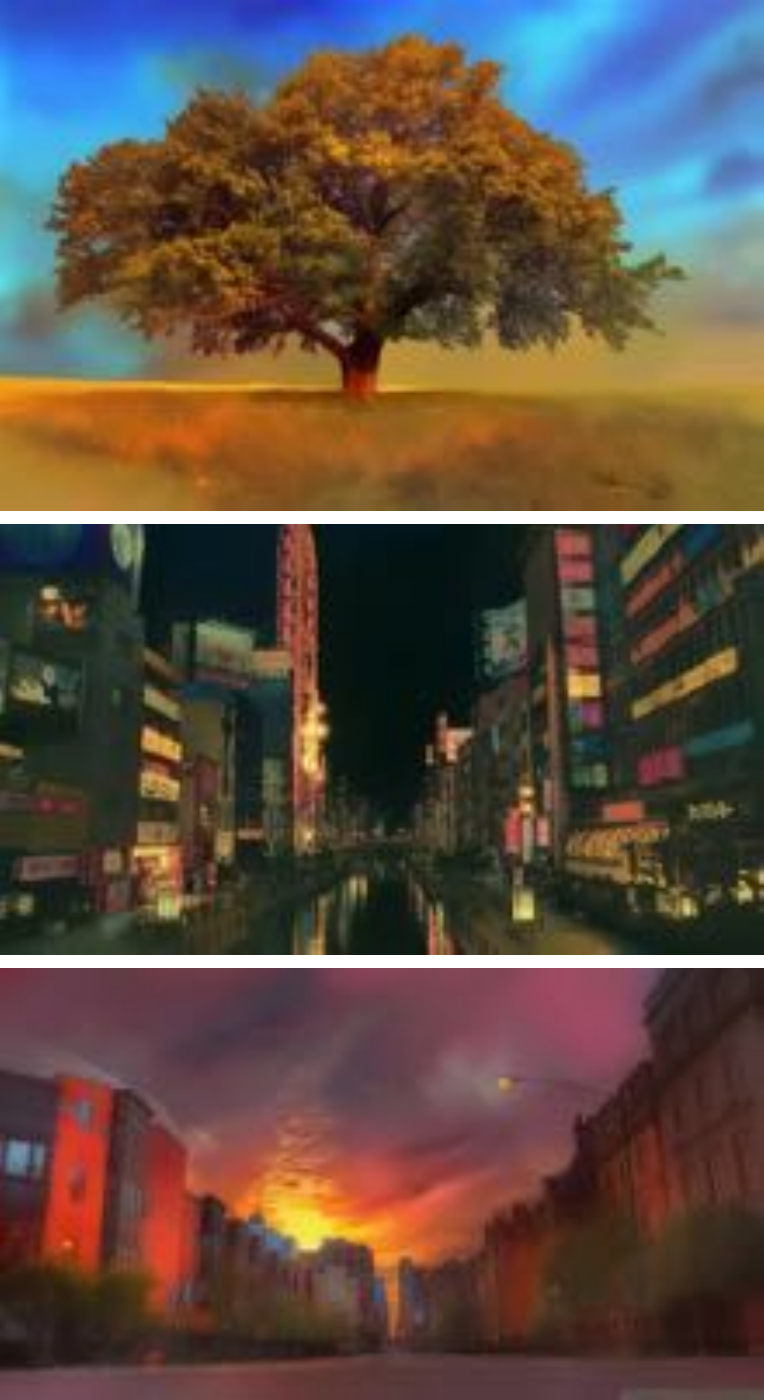}}\ 
    \subfloat[PhotoWCT-AE1]{\includegraphics[width=0.24\linewidth]{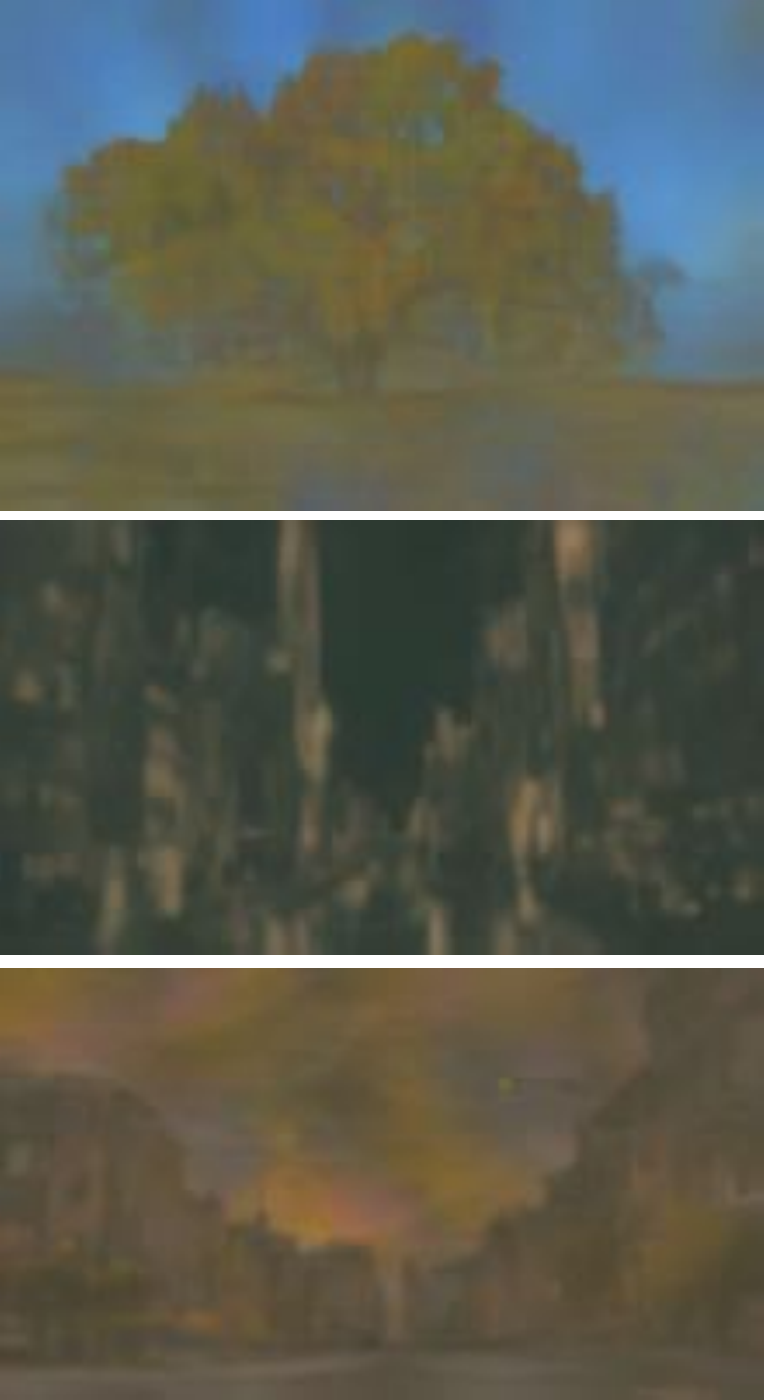}}\\
    \subfloat[StyleNAS-RS]{\includegraphics[width=0.24\linewidth]{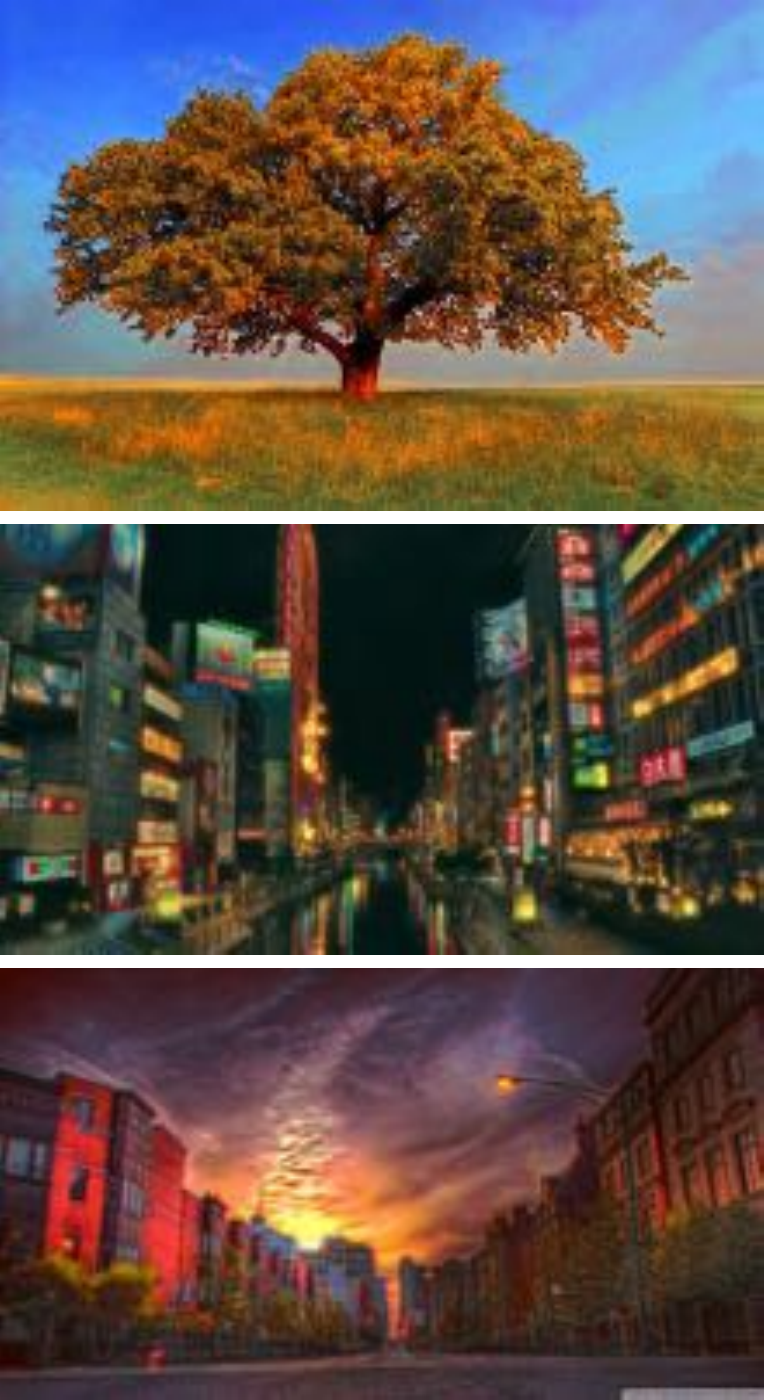}}\ 
    \subfloat[StyleNAS-5opt]{\includegraphics[width=0.24\linewidth]{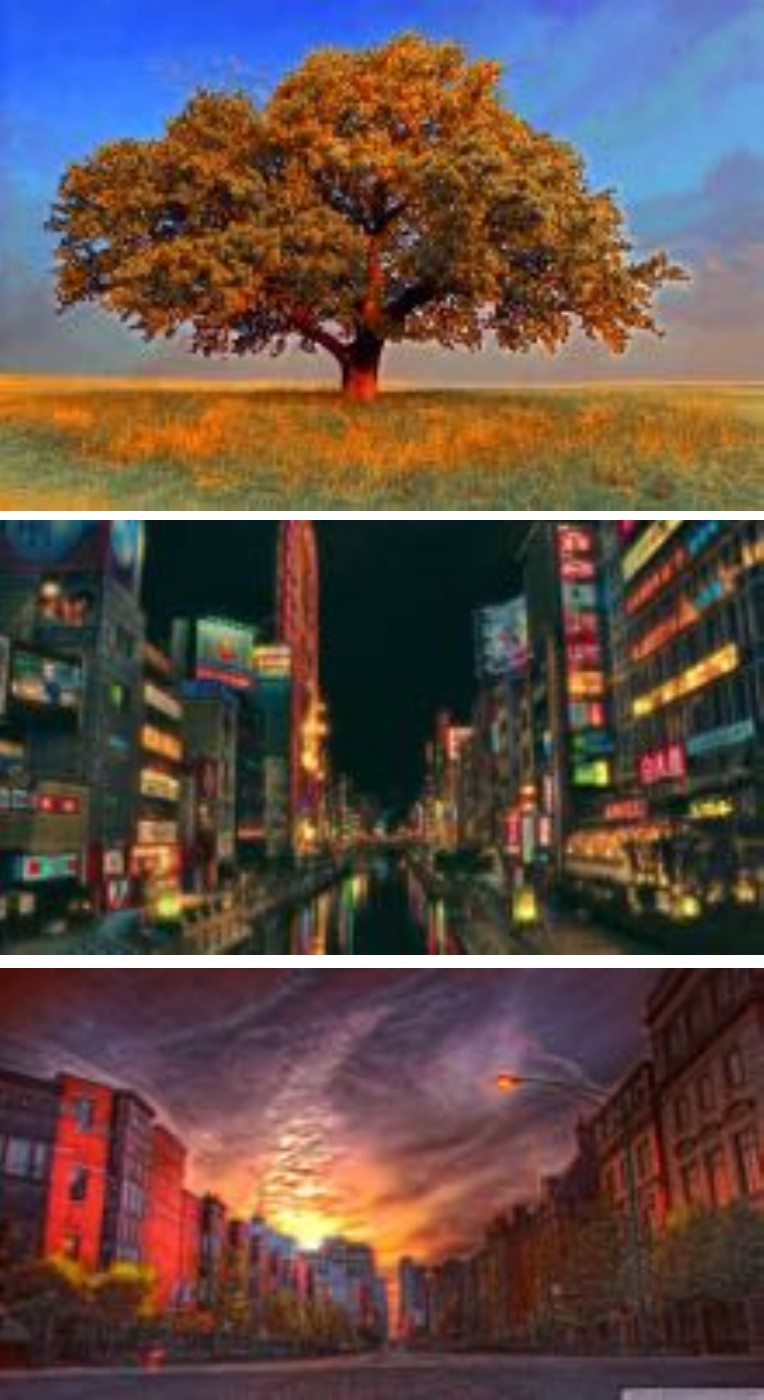}} \
    \subfloat[StyleNAS-7opt]{\includegraphics[width=0.24\linewidth]{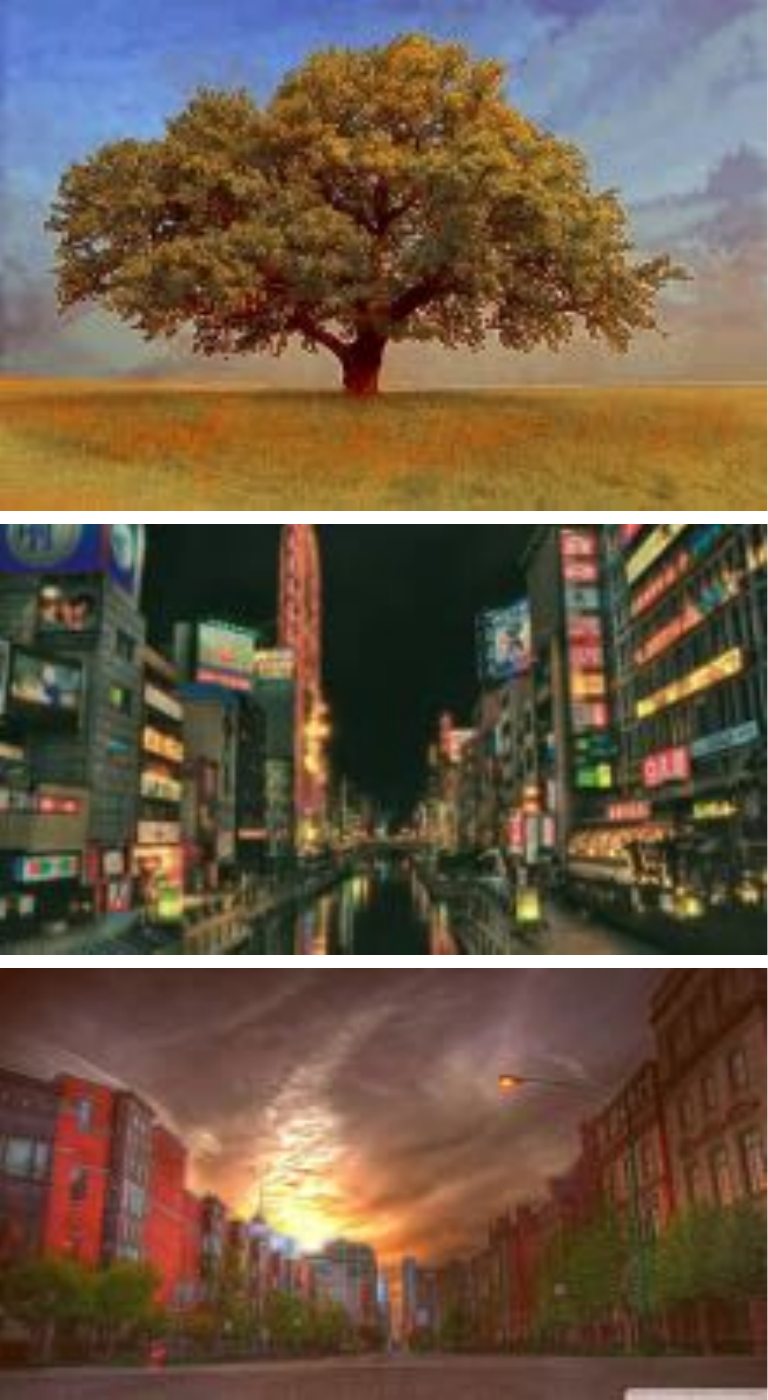}} \
    \subfloat[StyleNAS-9opt]{\includegraphics[width=0.24\linewidth]{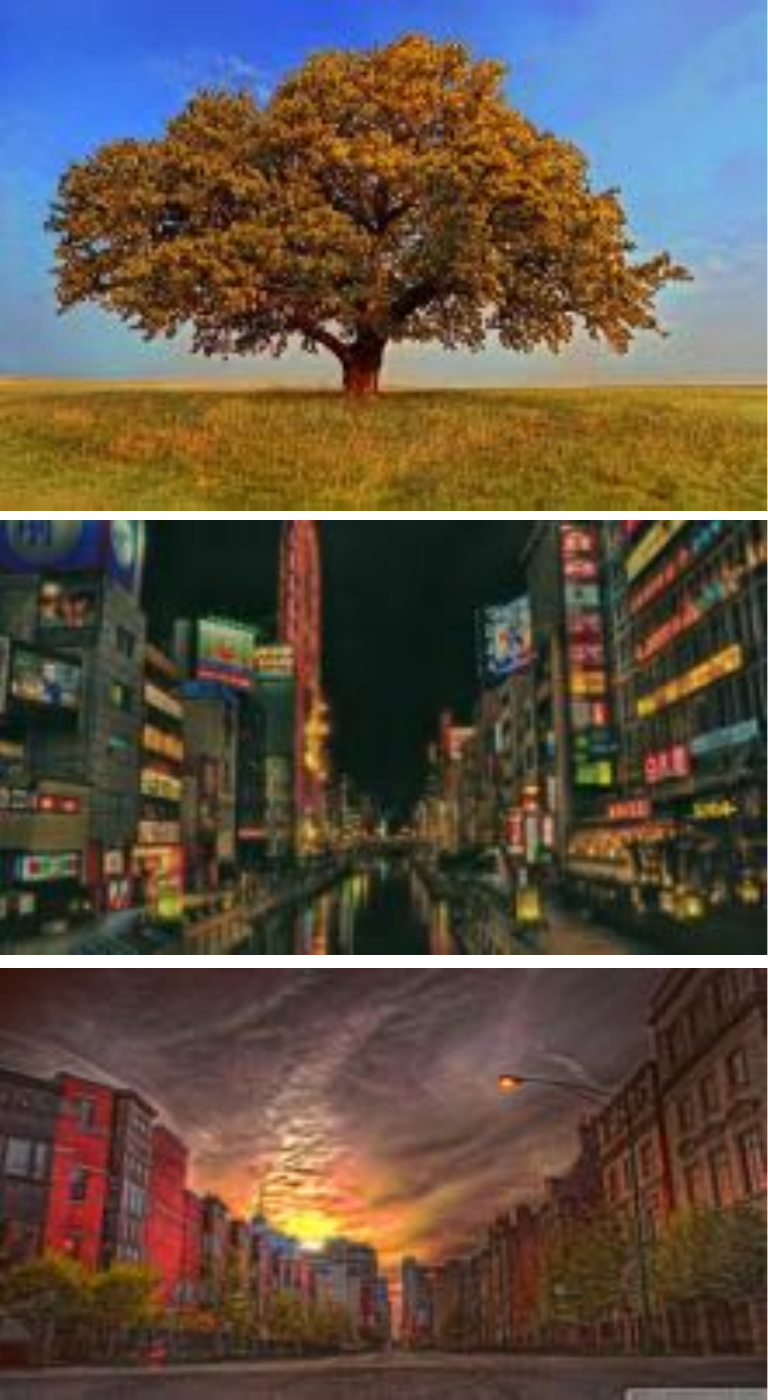}} 
\caption{\textbf{Photorealistic style transfer results comparison against state-of-the-art method and randomly searched architecture.} (\textbf{zoom-in} for details.)}
    \label{appendixfig:comp_photorealistic}
\end{figure*}

\subsection{More Searched Architectures by StyleNAS}
\begin{figure*}[t]

    \subfloat[Overall objective distribution of Random Search]{\includegraphics[width=0.48\textwidth]{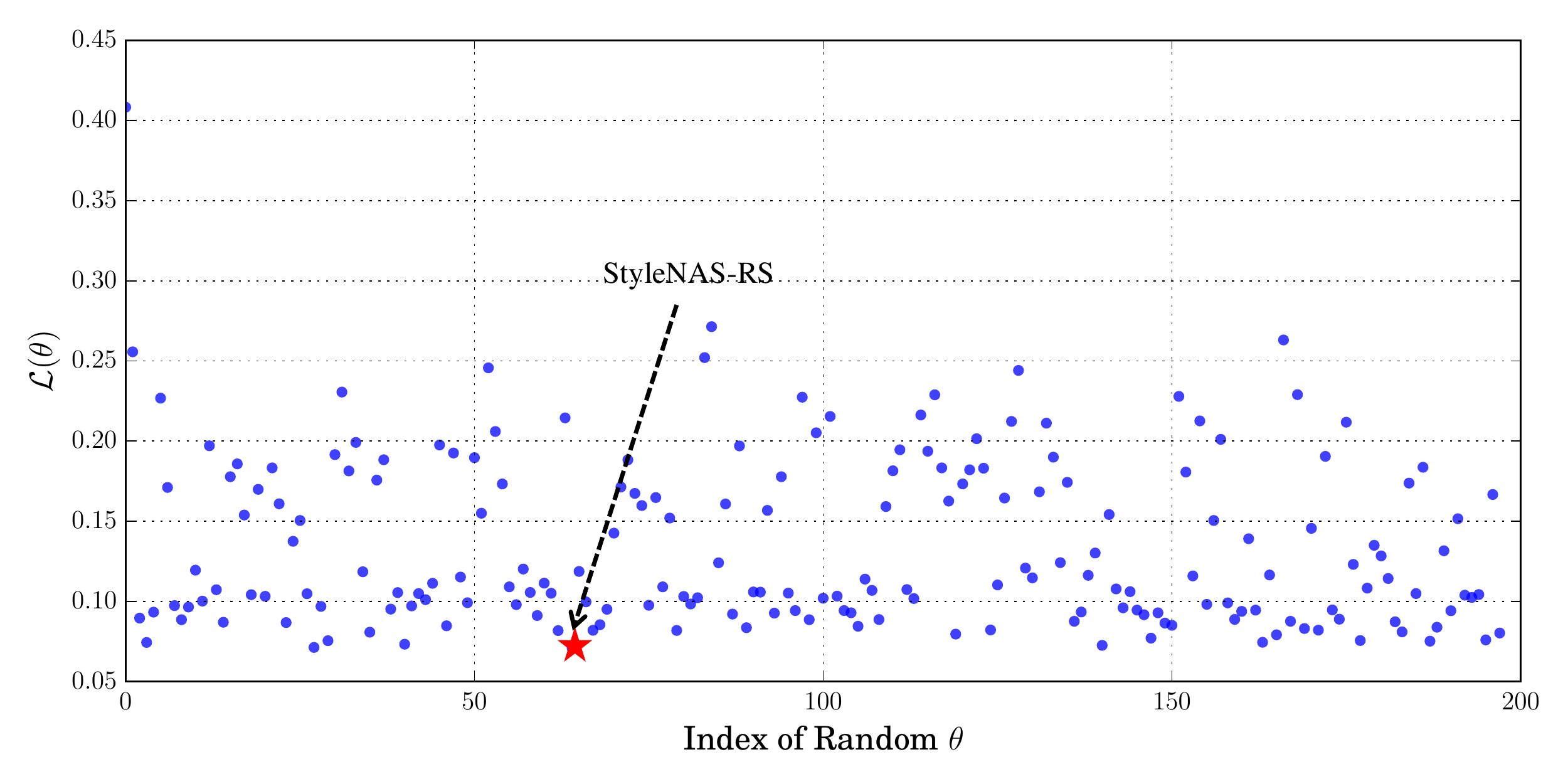}}\
    \subfloat[Overall objective decreasing trend of StyleNAS]{\includegraphics[width=0.48\textwidth]{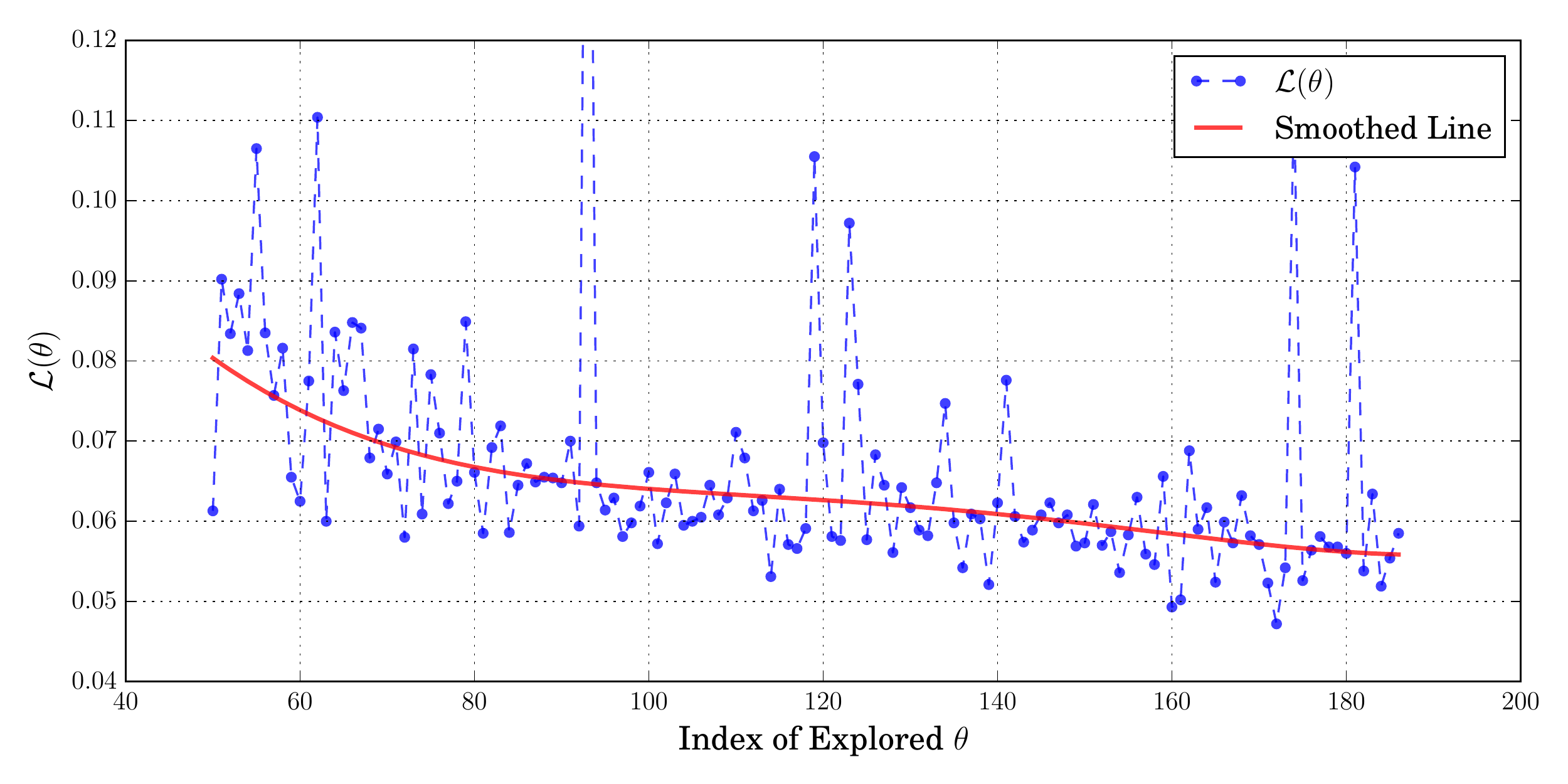}}\\
    \subfloat[Three objectives using StyleNAS]{\includegraphics[width=0.48\textwidth]{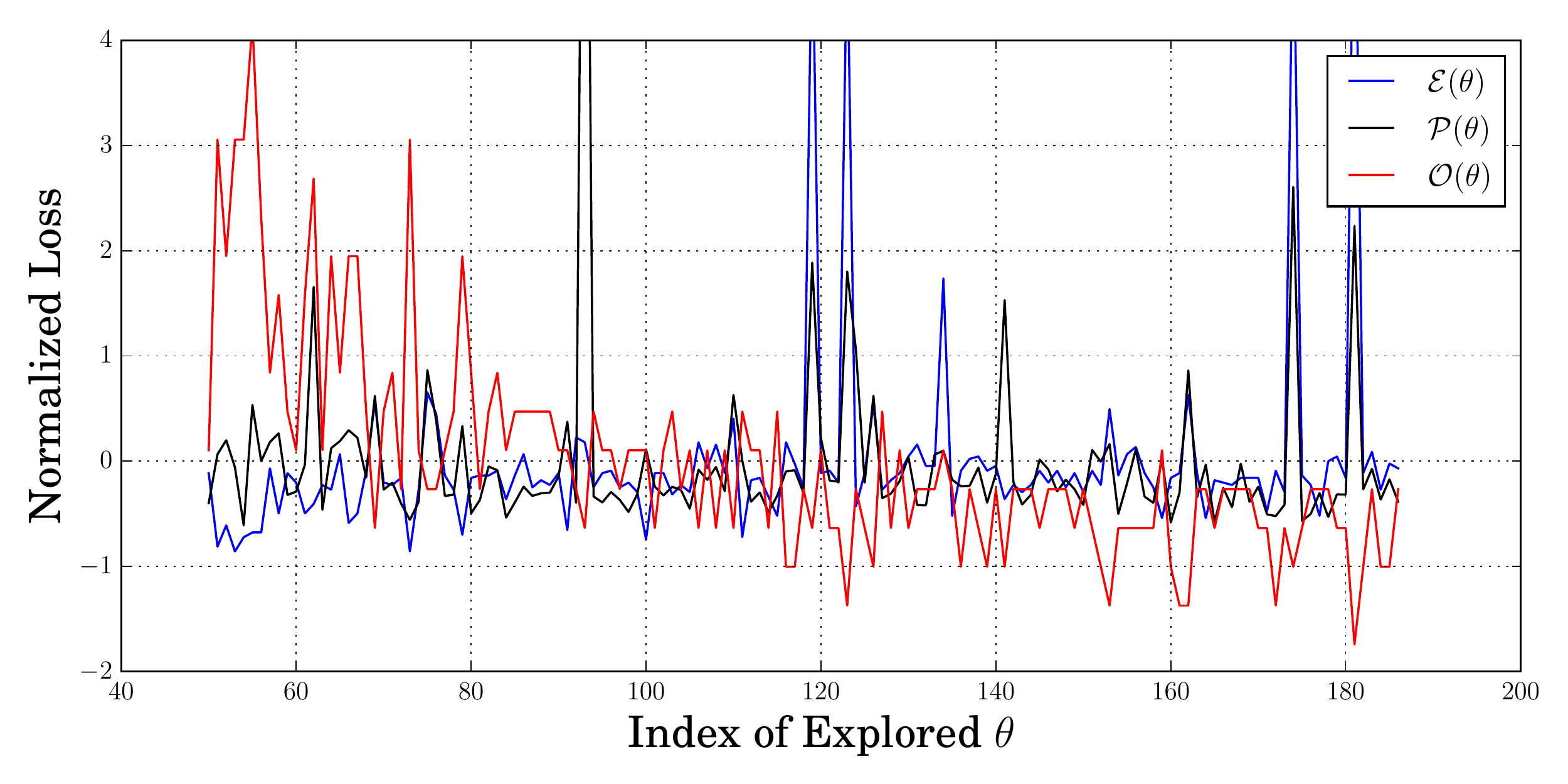}}\
    \subfloat[Hamming distance between $\theta$ and StyleNAS-7opt]{    \includegraphics[width=0.48\textwidth]{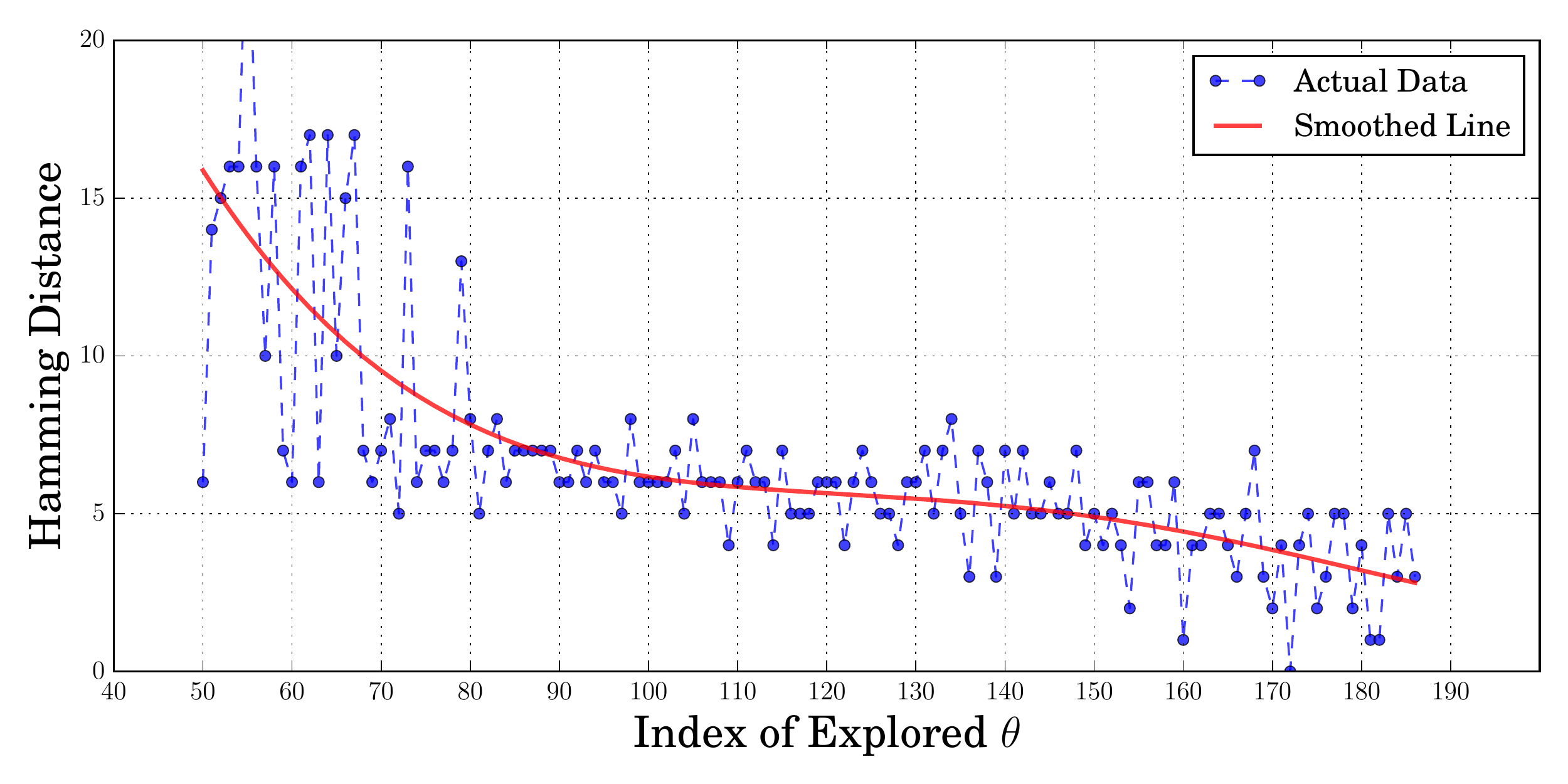}}\ 
    \caption{\textbf{Convergence of StyleNAS over index of explored architectures.}}
    \label{appendixfig:convergence}
\end{figure*}

The three decreasing lines shown in Figure~\ref{appendixfig:convergence}(c) demonstrate that the proposed StyleNAS algorithm can lower the percentage of operators used in the searched architecture $\theta$ while ensuring the image qualities i.e., $\mathcal{E}(\theta)$ and $\mathcal{P}(\theta)$ without any compromise.
We select three searched networks namely \emph{StyleNAS-5opt, StyleNAS-7opt, and StyleNAS-9opt} with 5, 7, and 9 operators among all operators which can be removed respectively. The architectures of these networks are shown in Figure~\ref{appendixfig:architecture}(b-d). Note that StyleNAS-7opt is the PhotoNAS in our paper.

As a reference, we provide two baseline algorithms to compare with StyleNAS:

\textbf{PhotoWCT-AE1.} A sub-architecture of PhotoWCT using the first auto-encoder (AE) of PhotoWCT~\cite{li2018closed} with the WCT modules placed at the bottleneck. The architecture is with a similar scale of parameters as StyleNAS networks.
    
\textbf{StyleNAS-RS.} A randomly searched network over the same search space of StyleNAS. We randomly draw and evaluates around $200$ architectures, where StyleNAS-RS is the best architecture with the lowest overall search objective.

Figure~\ref{appendixfig:comp_photorealistic} demonstrates the contrast examples of style transferred images produced by our three searched architectures (i.e., StyleNAS-5opt, StyleNAS-7opt, and StyleNAS-9opt searched by StyleNAS with the PhotoNet as supervisory oracle). We also compare the result with PhotoWCT~\cite{li2018closed}, PhotoWCT-AE1, and StyleNAS-RS. 
From our visual comparison, we observe that PhotoWCT~\cite{li2018closed} generates images with quite a lot of details lost (shown in (c)). For example, the grassland in the top image, the text on the advertising boards in the middle image, and the sky in the bottom image are blurred. 
Results of StyleNAS-RS (shown in (h)) have compromised stylization effects and the generated images are comparably of poor-quality. The StyleNAS-Xopt networks (shown in (e-g)) create images with abundant details without compromise of style transfer effects.
PhotoWCT-AE1 has a similar time-consumption as the searched models. However, the PhotoWCT-AE1 fails to generate photorealistic images, which demonstrates the strong ability of the StyleNAS in finding effective networks.

\subsection{Search Effectiveness Analysis of StyleNAS}
\begin{table*}[t]
    \caption{\textbf{Computation time comparison.}}
        \footnotesize
    \centering
    \begin{tabular}{lcccccc}
        \toprule
        Method \hspace{-2mm}&\hspace{-4mm} PhotoWCT \hspace{-2mm}& \hspace{-4mm}PhotoWCT-AE1 \hspace{-2mm}&\hspace{-4mm} StyleNAS-RS \hspace{-2mm}&\hspace{-4mm} StyleNAS-5opt \hspace{-2mm}&\hspace{-4mm} StyleNAS-7opt \hspace{-2mm}&\hspace{-4mm} StyleNAS-9opt \\
        \midrule
        $256\times128$ & 4.38 & 0.83 & 0.30 & \textbf{0.05} & 0.07 & 0.47\\
        $512\times256$ & 25.37& 0.99 & 0.35 & \textbf{0.09} & 0.10 & 0.67\\
        $768\times384$ & 64.73 & 1.10 & 0.42 & \textbf{0.15} & 0.18 & 0.76\\
        $1024\times512$ & 153.25 & - & 0.52 & \textbf{0.23} & 0.29 & 0.91\\
        \bottomrule
    \end{tabular}
\label{appendixtab:efficiency}
\end{table*}

Our experimental results show that StyleNAS is much more effective than Random Search. In our experiments, the StyleNAS algorithm explores 20 architectures per round. We let StyleNAS search architectures for 7 rounds. A total of 140 architectures were obtained. Among them, 137 architectures were evaluated and the rest 3 architectures failed in training. We also used the Random Search strategy to randomly draw 200 architectures from the search space. We picked up the best architectures from both search methods with the lowest overall objective. The best one from random search, i.e., StyleNAS-RS has the objective function value of 0.0709 while the best one from StyleNAS, i.e., StyleNAS-7opt has the objective function value of 0.0472 (in our study for the objective function value the smaller the better). 

We also perform quantitative analysis to compare architectures obtained by StyleNAS and Random Search. In terms of time consumption, StyleNAS-RS spends a significantly longer time than StyleNAS-7opt for the style transfer tasks of all image resolutions evaluated. While the fastest network searched (StyleNAS-5opt) only consumes 16\%$\sim$40\% time of StyleNAS-RS. Though StyleNAS-9opt spent longer time than StyleNAS-RS in our experiments. However, the image quality obtained by StyleNAS-9opt as well as StyleNAS-7opt is much better than StyleNAS-RS, which is demonstrated by Table~\ref{appendixtab:evaluation}. All in all, StyleNAS-RS cannot outperform StyleNAS-7opt in both quality and complexity wises. Please refer to Tables~\ref{appendixtab:efficiency} for detailed comparisons. Please note that the running time of StyleNAS-7opt, \ie, PhotoNAS is different from the one show in the body of the paper since these two batch of experiments are evaluated on different platforms.

\subsection{NAS Convergence Analysis}
StyleNAS can converge in terms of both overall search objective and the architectures. Figure~\ref{appendixfig:convergence}(b) shows that the overall search objective of StyleNAS decreases over the number of architectures explored. We further break down the overall objective to its three parts, where Figure~\ref{appendixfig:convergence}(c) demonstrates that while the two objectives of image qualities (i.e., reconstruction error and perceptual loss) are ensured at lower levels, the third objective --- the numbers of operators decrease in a general sense. The trends of searched networks of StyleNAS demonstrates the size of explored architectures became smaller and smaller while the stylization effects do not compromise.

We find the architectures searched by StyleNAS would also converge. We take the binary code of StyleNAS-7opt architecture as a reference, and estimate the hamming distance of every explored architecture in the \emph{history set} to the StyleNAS-7opt using binary codes. Figure~\ref{appendixfig:convergence}(d) provides the yet first evidence of the convergence in the search space, with simple evolutionary NAS strategies.

\section{Appendix D: User Study}
\begin{table*}[t]
    \caption{\textbf{User study result.}}
    \centering
   \begin{tabular}{lcccc}
        \toprule
        Method & DPST & PhotoWCT & WCT$^2$ & Ours(PhotoNAS) \\
        \midrule
        Preference Percentage$\uparrow$ & 23.89\% & 17.22\% & 16.11\% & \textbf{42.78\%}\\
        \bottomrule
    \end{tabular}
    \label{appendixtab:evaluation}
\end{table*}

We conduct a user study to subjectively demonstrate the effectiveness of the proposed PhotoNAS. We randomly select 25 content and style photo pairs to evaluate the photorealistic style transfer methods. We collect photorealistic stylization results of DPST~\cite{luan2017deep}, PhotoWCT~\cite{li2018closed}, WCT$^2$~\cite{yoo2019photorealistic}, and our PhotoNAS (\ie StyleNAS-7opt) without any pre- and post-processing for a fair comparison.
We design a webpage for users to vote at their best preferable results and for us to collect the statistical voting result. For each content and style pairs, we display results of DPST, PhotoWCT, WCT$^2$, and our PhotoNAS side-by-side in an anonymous random order and let the subject choose the best one in terms of less artifact, less distortion, and more details.
We collect a total of 180 votes.
The preference percentage of the choices are summarized in Table~\ref{appendixtab:evaluation}, which demonstrates that PhotoNAS improves over the results of DPST, PhotoWCT, and WCT$^2$ in terms of fewer artifacts, fewer distortions, and better detail preservation.

\section{Appendix E: Stylization Effects Control}
\begin{figure*}[t]
    \centering
    \includegraphics[width=1.0\textwidth]{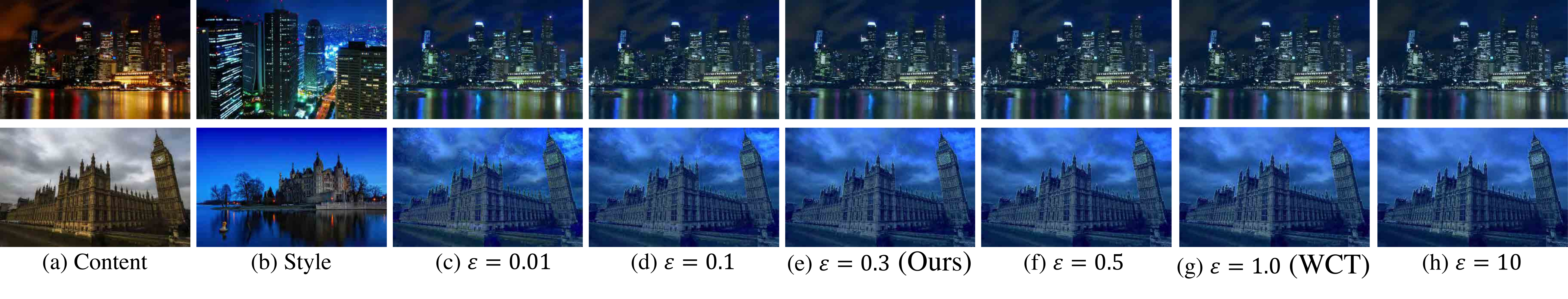}
    \caption{\textbf{Stylization effects with different $\varepsilon$ values.} The top row is an example that the style transfer effects remain intact while $\varepsilon$ changes. On the contrary, the bottom row shows an example that stylized images contain more noise with lower $\varepsilon$ values. Please zoom-in to see details of the sky in bottom images.}
    \label{appendixfig:8}
\end{figure*}

Traditional artistic style transfer methods using feed-forward auto-encoders (\eg, WCT, AdaIN) usually adopt the mixture of the transferred and the content features (get from VGG encoder) to control the effects of stylization. Our method, as an auto-encoder based approach, naturally support such a way to control style transfer effects. Moreover, as described in pre-analysis, architectural modules that impact photorealistic style transfer effects can also be used to fine-tune the style transfer effects according to user preference. Besides that, we find an inner-parameter of WCT has a subtle influence on the photorealistic stylization effects. We start to explain this by reviewing the detail of the WCT transfer module.
\begin{align}
    \hat{f_{cs}} = E_s D_s^{\frac{1}{2}} E_s^{T} \hat{f_c},
    \label{eq:1}
\end{align}

where

\begin{align}
    \hat{f_c} &= E_c D_c^{-\frac{1}{2}} E_c^{T} f_c \\
    f_c f_c^{T} &= E_c D_c E_c^T\\
    f_s f_s^T &= E_s D_s E_s^T,
\end{align}

so that we have

\begin{equation}
    \hat{f_{cs}} = E_s D_s^{\frac{1}{2}} E_s^{T} E_c D_c^{-\frac{1}{2}} E_c^{T} f_c
    \label{eq:2}
\end{equation}
Note that $f_c$ and $f_s$ denote zero-mean content and style features of the encoder respectively. $f_{cs} = \hat{f_{cs}} + m_s$ is the transferred feature, where $m_s$ denotes the mean of $f_s$. In Eq.~\ref{eq:2}, we have to compute $ D_c^{-\frac{1}{2}}$ in order to get $\hat{f_{cs}}$. To ensure the numerical stability, we should add a $\varepsilon$ to ensure every element in $D_c$ is not too close to zero. So the final equation to of WCT is

\begin{equation}
    \hat{f_{cs}} = E_s D_s^{\frac{1}{2}} E_s^{T} E_c \left( D_c + \varepsilon \right)^{-\frac{1}{2}} E_c^{T} f_c
    \label{eq:3}
\end{equation}

We find that this $\varepsilon$ have a subtle influence on the photorealistic stylization effects. Fig.~\ref{appendixfig:8} shows photorealistic style transfer results with a sequence of $\varepsilon$. We find that an over-small $\varepsilon$ usually make the produced result contains too much noise if the content photo is noisy. Please see the bottom row in Fig.~\ref{appendixfig:8} for example of a noisy result caused by too small $\varepsilon$. Surprisingly, in some cases, the stylized effects remain intact with a small or a comparably large $\varepsilon$. Please see the top row of Fig.~\ref{appendixfig:8} for example. We think these two controversial cases demonstrate that $E_s$ and $E_c$ in Eq.~\ref{eq:3} have stronger influence against $D_c$ and $D_s$. Overall, $\varepsilon$ plays a role to balance signal-noise ratio and content preservation in some cases. We choose $\varepsilon=0.3$ in our experiment. Please note that WCT uses $\varepsilon=1$, which is expected to have better content details preservation since they used a higher $\varepsilon$. But our method preserves much more fine details, which demonstrates the effectiveness of our algorithm. Moreover, We allow users to adjust $\varepsilon$ to get most preferable stylization results.

\section{Appendix F: Photorealistic Video Transfer}
\begin{figure*}[t]
    \centering
    \includegraphics[width=1.0\textwidth]{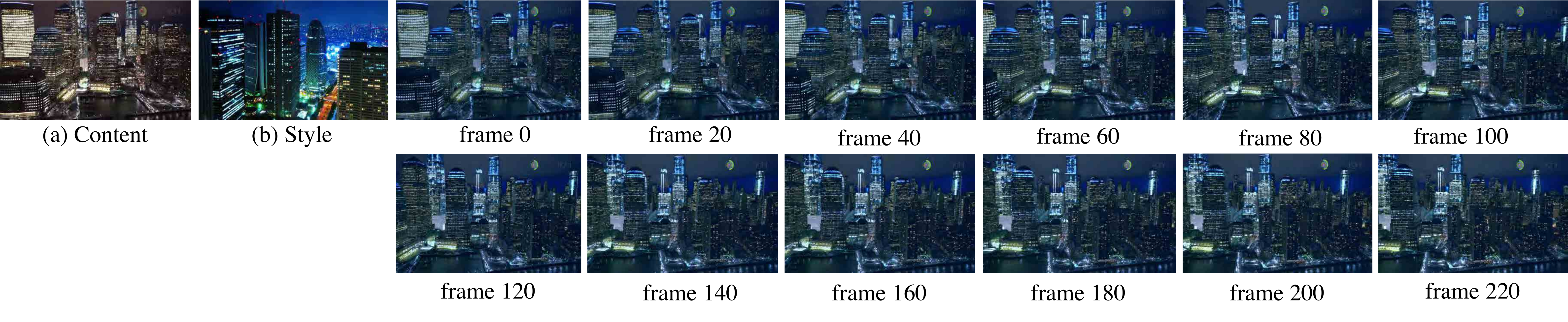}
    \caption{\textbf{Photorealistic video transfer results.}}
    \label{appendixfig:9}
\end{figure*}
Since our algorithm has a strong ability to preserve content details. the proposed method can be directly used to make photorealistic video transfer without any specific modification. We show a sequence of video frames in Fig.~\ref{appendixfig:9}. The transferred frames have almost the same style and the video is stable. You may find the transferred video in our uploaded files.

\section{Appendix G: Our Failure Cases}
\begin{figure*}[t]
    \centering
    \includegraphics[width=\textwidth]{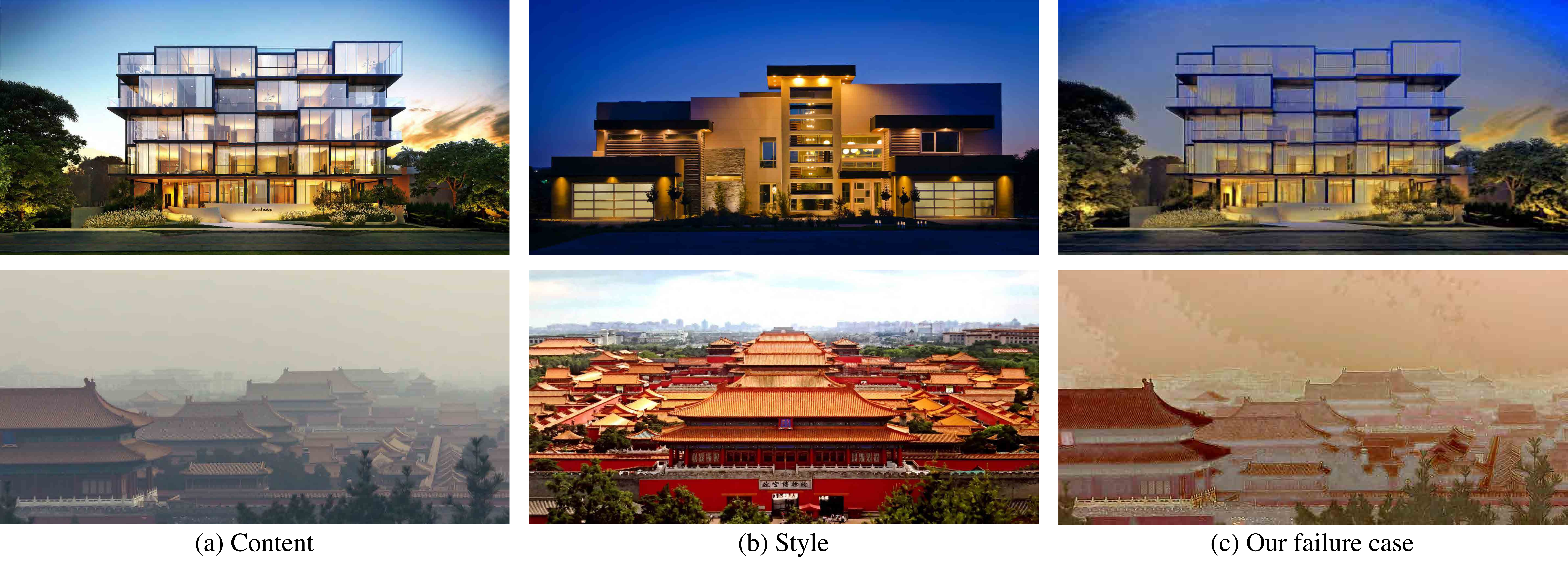}
    \caption{\textbf{Our failure cases.}}
    \label{appendixfig:10}
\end{figure*}

Due to that we do not use pre-conditioned regional masks and smooth-oriented post-processing, our method may produce two kinds of failure cases: 1) If the style photo contains several remarkably different style factors, our method may fade in these cases. Please see the top row in Fig.~\ref{appendixfig:10} (c) for example. The upper part of the building is rendered according to the style of the blue sky rather than the yellow light. 2) If the input content photo contains significant noise, the produced image may also look noisy. Such a noisy image may look non-photorealistic in certain styles. We show an example of this case in the bottom row of Fig.~\ref{appendixfig:10} (c). We will try to fix these failure cases in our future work.

\small
\bibliographystyle{aaai}
\bibliography{main}
\clearpage

\end{document}